\pdfoutput=1

\documentclass[11pt]{article}

\usepackage[final]{acl}

\usepackage{times}
\usepackage{latexsym}

\usepackage[T1]{fontenc}

\usepackage[utf8]{inputenc}

\usepackage{microtype}

\usepackage{inconsolata}

\usepackage{graphicx}
\usepackage{amsmath}
\usepackage{amssymb}
\usepackage{mathtools}
\usepackage{amsthm}
\usepackage{hyperref}
\usepackage{amssymb}
\usepackage{multirow}
\usepackage{xcolor}
\usepackage{colortbl}
\usepackage{booktabs}

\usepackage{algorithm}
\usepackage{algorithmicx}
\usepackage{algpseudocode}
\usepackage{verbatim}
\usepackage{caption}
\usepackage{subcaption}
\usepackage{fancyvrb}
\usepackage{fvextra}
\usepackage[most]{tcolorbox}

\definecolor{GREEN}{HTML}{62b197}
\definecolor{RED}{HTML}{e18e6d}

\DeclareUnicodeCharacter{FF5C}{\textbar}

\title{\centering MulDimIF: A Multi-Dimensional Constraint Framework for Evaluating and Improving Instruction Following in Large Language Models}

\author{
    \bf{\normalsize
    Junjie Ye$^{1}$\thanks{Equal contributions.}, Caishuang Huang$^{1*}$, Zhuohan Chen$^{1}$, Wenjie Fu$^{1}$,}\\ 
    \bf{\normalsize  Chenyuan Yang$^{1}$, Leyi Yang$^{1}$, Yilong Wu$^{1}$, Peng Wang$^{2}$, Meng Zhou$^{3}$,}\\
    \bf{\normalsize  Xiaolong Yang$^{3}$, Tao Gui$^{1}$, Qi Zhang$^{1,4,5}$\thanks{Corresponding author.}, Zhongchao Shi$^{2}$,  Jianping Fan$^{2}$, Xuanjing Huang$^{1}$}\vspace{2mm}\\ 
  {$^1$\normalsize Fudan University \ \ $^2$\normalsize Lenovo Research  \ \ $^3$\normalsize Tencent}\\
  {$^4$\normalsize Shanghai Artificial Intelligence Laboratory}\\ {$^5$\normalsize Shanghai Key Lab of Intelligent Information Processing}\vspace{2mm}\\
  \texttt{\normalsize jjye23@m.fudan.edu.cn, qz@fudan.edu.cn}\\
  }

\begin{document}
\maketitle
\begin{abstract}

Instruction following refers to the ability of large language models (LLMs) to generate outputs that satisfy all specified constraints.  Existing research has primarily focused on constraint categories, offering limited evaluation dimensions and little guidance for improving instruction-following abilities.
To address this gap, we introduce~\emph{MulDimIF}, a multi-dimensional constraint framework encompassing three constraint patterns, four constraint categories, and four difficulty levels. Based on this framework, we design a controllable instruction generation pipeline. Through constraint expansion, conflict detection, and instruction rewriting, we construct 9,106 code-verifiable samples.
We evaluate 18 LLMs from six model families and find marked performance differences across constraint settings. For instance, average accuracy decreases from 80.82\% at Level I to 36.76\% at Level IV.
Moreover, training with data generated by our framework significantly improves instruction following without compromising general performance. In-depth analysis indicates that these gains stem largely from parameter updates in attention modules, which strengthen constraint recognition and adherence.
Code and data are available in~\url{https://github.com/Junjie-Ye/MulDimIF}.
\end{abstract}

\section{Introduction}

\begin{figure}[!t]
    \centering
    \includegraphics[width=\linewidth]{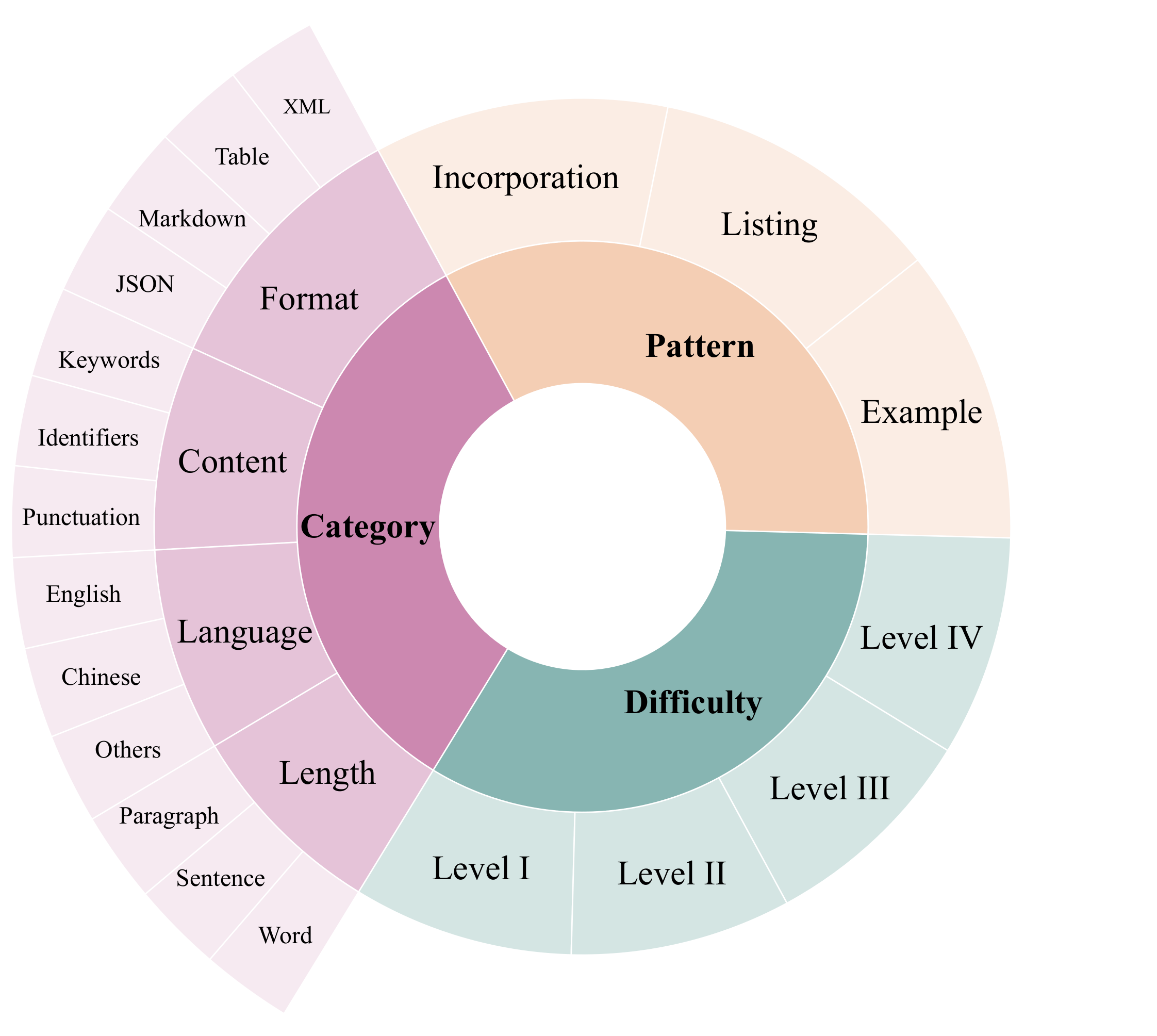}
    \caption{The hierarchical structure of the multi-dimensional constraint framework, which includes three constraint patterns, four constraint categories (subdivided into thirteen subcategories), and four levels of constraint difficulty.}
    \label{fig:framework}
\end{figure}

Instruction following is a fundamental capability of large language models (LLMs)~\cite{Claude, GPT-4, Gemini1.5, Qwen2.5}, enabling them to generate outputs that satisfy user-specified constraints~\cite{IF-Bench, IF-RAG, LLM-Eval3}. This skill is essential in real-world applications, especially in agentic and tool-assisted workflows where outputs must adhere to strict formats such as JSON~\cite{agent-survey, IF-Agents, TL-Training}. Even small deviations can cause parsing failures and downstream system breakdowns~\cite{ToolEyes}.

Existing research has approached this challenge along two main lines. Benchmarking efforts typically rely on template-based instructions~\cite{IFEval, Multi-IF}, evaluated either through code verification or LLM-as-a-judge methods~\cite{FollowBench}. In parallel, training-focused studies have explored improving instruction following via tree search~\cite{SPaR}, reinforcement learning (RL)~\cite{C2S}, and other data-centric strategies.

However, these efforts face key limitations. Benchmarking work largely emphasizes the diversity of constraint categories~\cite{IFEval, Conifer, InfoBench, IF-Bench}, while relying on narrow evaluation dimensions that cannot fully characterize instruction-following ability. Meanwhile, training approaches often boost benchmark scores through data engineering~\cite{IOPO, SPaR, Self-Play}, but rarely examine the mechanisms driving these improvements. As a result, we still lack deeper insight into the factors that shape instruction-following performance.

To address this gap, we present~\emph{MulDimIF}, a multi-dimensional constraint framework for evaluating and improving instruction following in LLMs. As shown in Figure~\ref{fig:framework}, the framework introduces three constraint patterns, namely example, listing, and incorporation, which capture common user instruction formats. Constraints are organized into four major categories, namely content, language, format, and length, which further divide into thirteen subcategories. The framework also defines four difficulty levels based on the complexity of constraint combinations within instructions.
Building on this framework, we design an instruction generation pipeline consisting of constraint expansion, conflict detection, and instruction rewriting. This pipeline enables controlled transformation of initial instructions into diverse, constraint-rich variants. Using it, we generate 9,106 code-verifiable data instances.

We evaluate 18 LLMs across six model families and uncover significant variation in their ability to follow different forms of constraints. Notably, while models perform well on the example pattern, they struggle with listing and incorporation patterns, emphasizing the challenges posed by complex instructions and the benefits of few-shot prompting~\cite{GPT-3}. On average, performance declines from 80.82\% at Level I to 36.76\% at Level IV, with even the best model scoring only 67.50\% overall.

To leverage these findings, we train LLMs using the GRPO algorithm~\cite{DeepSeekMath} on data generated by our framework. The trained models show significantly improved instruction-following abilities without sacrificing general performance. Parameter-level analysis and case studies indicate that these gains primarily result from updates within attention modules, which more effectively align models' focus with specified constraints.

Our contributions are summarized as follows:
\begin{itemize}
    \item We propose a multi-dimensional constraint framework that evaluates and improves instruction following in LLMs;
    \item We design a controllable instruction generation pipeline that transforms plain instructions into constraint-rich variants;
    \item We generate 9,106 diverse constraint-rich data instances, evaluate the instruction-following abilities of 18 LLMs, and improve model performance through targeted training;
    \item We conduct parameter-level analysis, showing that improvements in instruction following primarily arise from updates within attention modules.
\end{itemize}

\begin{figure*}[!t]
    \centering
    \includegraphics[width=\linewidth]{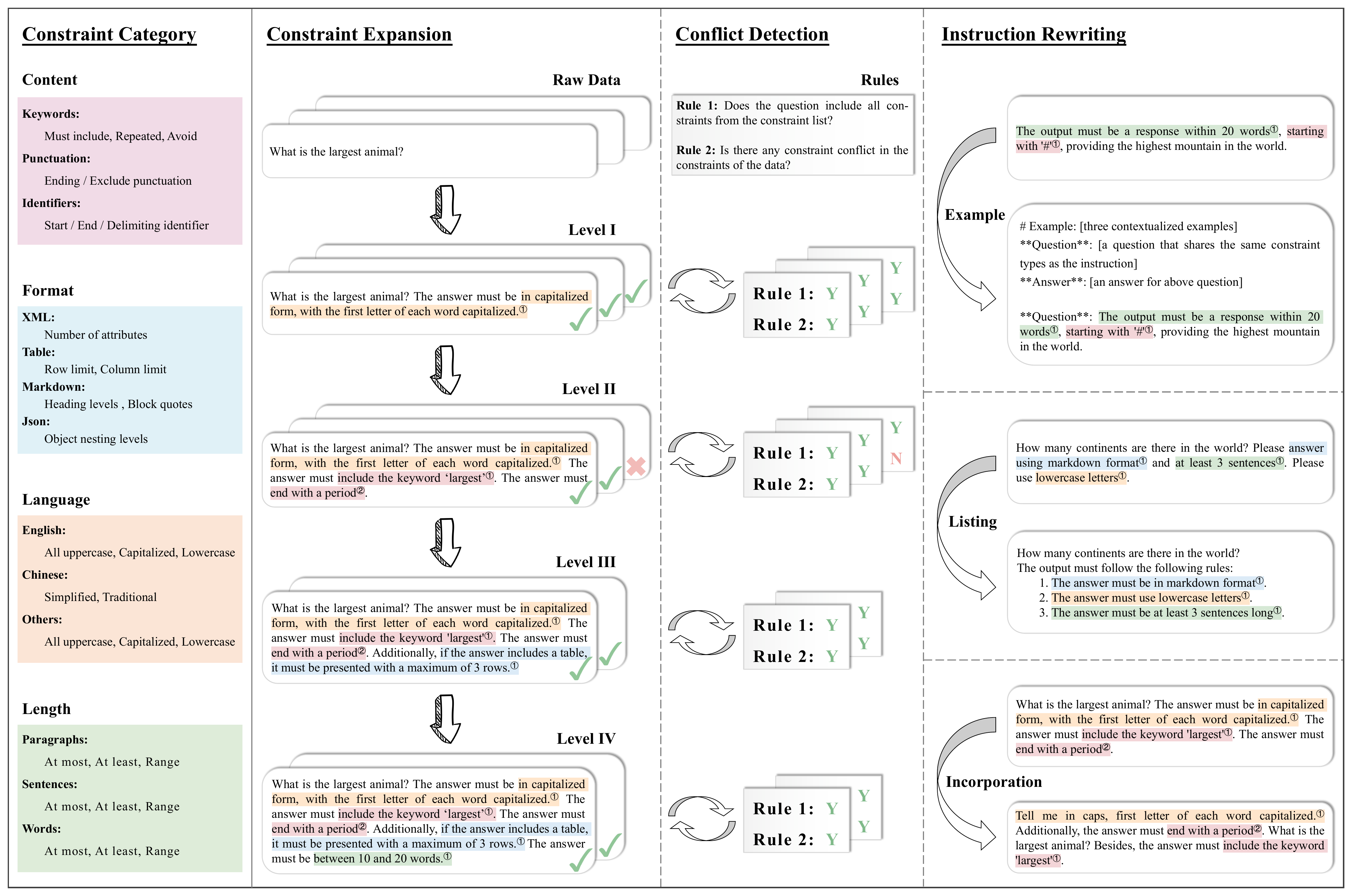}
    \caption{Illustration of the instruction generation pipeline. \textbf{Constraint Expansion}: Randomly selects a constraint category not yet included in the instruction and adds 1–2 specific constraints. \textbf{Conflict Detection}: Identifies whether the new instruction introduces redundant constraints or conflicts, and discards conflicting instructions. \textbf{Instruction Rewriting}: Rewrites the remaining instructions based on different constraint patterns.}
    \label{fig:pipeline}
\end{figure*}

\section{Related Works}

\paragraph{Evaluation of Instruction Following}
Evaluating the instruction-following abilities of LLMs has become a central focus in recent research~\cite{LLM-Eval}. Benchmarks such as IFEval~\cite{IFEval}, FollowEval~\cite{FollowEval}, and FollowBench~\cite{FollowBench} evaluate models on dimensions like logical reasoning and stylistic consistency, using either code-based or LLM-based evaluations. Multi-IF~\cite{Multi-IF} extends this to multilingual, multi-turn dialogue settings, while InfoBench~\cite{InfoBench} decomposes complex instructions into simpler subtasks to evaluate execution accuracy. CIF-Bench~\cite{CIFBench} focuses on the generalization abilities of Chinese LLMs under zero-shot scenarios. Despite their breadth, many of these benchmarks rely on templated or highly constrained prompts, which limits their ability to capture real-world instruction diversity and support fine-grained evaluation. Our work addresses these limitations by introducing a multi-dimensional constraint framework comprising three constraint patterns, four constraint categories, and four difficulty levels. Built on this framework, we develop a controllable instruction generation pipeline that enhances diversity and complexity through constraint expansion, conflict detection, and instruction rewriting.

\paragraph{Improving of Instruction Following}
A range of algorithms have been proposed to improve the instruction-following performance of LLMs. RL approaches such as PPO~\cite{PPO} and DPO~\cite{DPO} optimize model behavior based on user preferences. IOPO~\cite{IOPO} augments this by aggregating question-answer pairs across datasets to enrich preference signals and refine the optimization objective. Coninfer~\cite{Conifer} adopts a curriculum learning approach, incrementally increasing task difficulty during fine-tuning to improve constraint handling. While these methods yield measurable gains, they often lack in-depth analysis of the model characteristics driving these improvements, which limits their interpretability and generalizability. To fill this gap, we introduce a comprehensive data construction pipeline, enabling the creation of high-quality instruction-following datasets. Our parameter-level analysis suggests that a significant portion of the performance improvement arises from tuning the model's attention mechanisms, which enhances its ability to recognize and comply with constraints.

\section{Approaches}

\begin{table*}[!t]
    \centering
    \resizebox{\linewidth}{!}
    {
    \begin{tabular}{lcccccccccccc}
    \toprule
     \multirow{2}*{\textbf{Data}} &  \multicolumn{3}{c}{\textbf{Constraint Pattern}} & \multicolumn{4}{c}{\textbf{Constraint Category}} & \multicolumn{4}{c}{\textbf{Constraint Difficulty}} & \multirow{2}*{\textbf{Total}}    \\ \cmidrule(lr){2-4} \cmidrule(lr){5-8} \cmidrule(lr){9-12}
         & \textbf{Example} & \textbf{Listing} & \textbf{Incorporation} & \textbf{Content} & \textbf{Format} & \textbf{Language} & \textbf{Length} & \textbf{Level I} & \textbf{Level II} & \textbf{Level III} & \textbf{Level IV} \\ \midrule
         Training & 1225 & 3308 & 3373 & 8888 & 9850 & 6168 & 9541 & 610 & 1614 & 2522 & 3160 & 7906\\
         Test & 400 & 400 & 400 & 1175 & 1210 & 694 & 1101 & 300 & 300 & 300 & 300 & 1200\\
         \bottomrule
    \end{tabular}
    }
    \caption{Distribution of training and test data constructed on the automated instruction generation pipeline.}
    \label{tab:data}
\end{table*}


\subsection{Multi-Dimensional Constraint Framework} 
\label{sec:framework} 
Existing studies~\cite{IFEval, FollowEval, FollowBench} often focus solely on the diversity of constraint categories, limiting the dimension for evaluating and improving the instruction-following ability. To address this limitation, we propose a multi-dimensional constraint framework, as shown in Figure~\ref{fig:framework}.\footnote{Examples are available in Appendix~\ref{sec:examples}.}

\subsubsection{Constraint Pattern} 
Drawing inspiration from publicly available guidelines for writing instructions~\cite{prompt_engeer}, we identify three common patterns used to introduce constraints during interactions with LLMs.

\paragraph{Example}
The example pattern involves adding several question-answer pairs that share the same constraint type as the instruction to be followed. This method enhances the model's ability to comply with constraints through contextualized examples, a technique commonly known as in-context learning~\cite{GPT-3}.

\paragraph{Listing}
The listing pattern presents constraints in a clearly structured, point-by-point format. This approach provides explicit communication of constraint requirements, making it especially effective in zero-shot scenarios.

\paragraph{Incorporation} 
The incorporation pattern integrates constraints directly into the instruction, rather than listing them separately. While this approach maintains fluency, it may make it more difficult for LLMs to clearly interpret individual constraint requirements.

\subsubsection{Constraint Category} 
Beyond the method of presentation, the types of constraints can vary widely. To enable fine-grained analysis, we categorize constraints into four main categories with thirteen subcategories.

\paragraph{Content} 
Content constraints restrict the elements present in the model's output. These can be further divided into three subcategories: ensuring the inclusion of specific keywords, adhering to particular punctuation requirements, or referencing predefined identifiers.

\paragraph{Format} 
Format constraints require the output to follow specific structural rules, often necessary for post-processing tasks. Common examples include outputs in XML, Table, Markdown, or JSON.

\paragraph{Language} 
Language constraints specify the language to be used in the output, which is fundamental in translation tasks~\cite{analy-ye}. Based on the language category, constraints are classified into English, Chinese, or other languages.

\paragraph{Length} 
Length constraints enforce limits on the output's size. Depending on granularity, these constraints can apply at the paragraph level, sentence level, or word level.

\subsubsection{Constraint Difficulty} 
In addition to presentation and type, the number of constraints also impacts task difficulty. We define four difficulty levels based on the number and variety of constraints present in the instruction.

\paragraph{Level I}
Level I includes an instruction with a single type of constraint, containing one or two individual constraint elements.

\paragraph{Level II}
Level II includes an instruction with two types of constraints, comprising a total of two to four individual constraint elements.

\paragraph{Level III}
Level III includes an instruction with three types of constraints, comprising a total of three to six individual constraint elements.

\paragraph{Level IV}
Level IV includes an instruction with four types of constraints, comprising a total of four to eight individual constraint elements.

\begin{table*}[!t]
    \centering
    \resizebox{\linewidth}{!}
    {
    \begin{tabular}{llcccccccccccc}
    \toprule
    \multirow{2}*{\textbf{Family}} & \multirow{2}*{\textbf{Version}} & \multicolumn{3}{c}{\textbf{Constraint Pattern}} & \multicolumn{4}{c}{\textbf{Constraint Category}} & \multicolumn{4}{c}{\textbf{Constraint Difficulty}} & \multirow{2}*{\textbf{Overall}}    \\ \cmidrule(lr){3-5} \cmidrule(lr){6-9} \cmidrule(lr){10-13}
        & & \textbf{Example} & \textbf{Listing} & \textbf{Incorporation} & \textbf{Content} & \textbf{Format} & \textbf{Language} & \textbf{Length} & \textbf{Level I} & \textbf{Level II} & \textbf{Level III} & \textbf{Level IV} & \\ \midrule
        \rowcolor{gray!10} \multicolumn{14}{c}{\textit{Open-Source LLMs (Non Reasoning)}} \\ 
    \multirow{2}*{LLaMA3.1} & Instruct-8B & 40.25 & 36.00 & 32.25 & 80.13 & 64.74 & 44.28 & 49.60 & 64.67 & 40.67 & 27.33 & 12.00 & 36.17\\
    & Instruct-70B & 68.00 & 54.25 & 48.25 & 86.62 & 73.65 & 80.90 & 65.73 & 78.00 & 61.33 & 51.33 & 36.67 & 56.83\\ \midrule
    \multirow{3}*{Qwen3} 
    & 8B-Direct.  & 63.00 & 55.00 & 51.00 & 84.03 & 72.27 & 83.00 & 69.89 & 87.67 & 60.33 & 45.67 & 31.00 & 56.17\\
    & 14B-Direct. & 64.00 & 63.00 & 52.00 & 87.92 & 77.00 & 89.87 & 65.86 & 88.00 & 66.67 & 50.67 & 33.67 & 60.00\\
    & 32B-Direct. & 70.00 & 59.00 & 49.00 & 87.14 & 78.29 & 78.29 & 68.01 & 85.00 & 65.67 & 51.00 & 34.00 & 58.92\\ \midrule
        \rowcolor{gray!10} \multicolumn{14}{c}{\textit{Open-Source LLMs (Reasoning)}} \\ 
    DeepSeek-R1- & Instruct-8B & 42.00 & 28.00 & 28.00 & 80.00 & 50.44 & 44.72 & 53.09 & 59.67 & 40.33 & 18.33 & 12.33 & 32.67\\
    Distill-LLaMA & Instruct-70B & 59.50 & 64.00 & 57.50 & 84.55 & 83.69 & 84.80 & 65.32 & 77.00 & 63.00 & 57.00 & 44.33 & 60.33\\ \midrule
    \multirow{3}*{Qwen3} 
    & 8B-Reason.  & 67.75 & 61.25 & 54.00 & 85.84 & 71.02 & 87.70 & 72.58 & 87.00 & 68.33 & 52.67 & 36.00 & 61.00\\
    & 14B-Reason. & 71.50 & 64.00 & 57.75 & 87.79 & 77.29 & \textbf{92.47} & 72.31 & 86.00 & 72.33 & 58.67 & 40.67 & 64.42\\
    & 32B-Reason. & 70.50 & 69.50 & 59.50 & 87.14 & 83.06 & 90.30 & 74.19 & 86.33 & 72.67 & 61.00 & 46.00 & 66.50\\ \midrule
        \rowcolor{gray!10} \multicolumn{14}{c}{\textit{Closed-Source LLMs}} \\ 
    \multirow{2}*{Gemini1.5} & Flash & 71.50 & 61.00 & 64.50 & 88.70 & 80.68 & 87.84 & 74.06 & 86.00 & 68.00 & 57.67 & 51.00 & 65.67\\
    & Pro & \textbf{73.50} & 61.75 & \textbf{65.25} & 87.40 & 81.81 & 88.28 & 75.67 & \textbf{86.67} & 70.00 & 59.00 & 51.67 & 66.83\\ \midrule
    \multirow{2}*{Claude3.5} & Haiku & 44.25 & 53.50 & 52.00 & 84.29 & 82.81 & 49.06 & 68.01 & 71.33 & 51.67 & 41.67 & 35.00 & 49.92\\
    & Sonnet & 72.50 & \textbf{69.00} & 61.00 & 86.62 & 83.44 & 84.23 & \textbf{76.21} & 82.67 & \textbf{71.67} & \textbf{60.67} & \textbf{55.00} & \textbf{67.50}\\ \midrule
    \multirow{4}*{GPT} & 3.5-Turbo & 49.00 & 41.25 & 38.00 & 84.94 & 70.39 & 42.98 & 66.26 & 77.33 & 48.00 & 30.33 & 15.33 & 42.75\\
    & 4-Turbo & 70.75 & 59.25 & 52.25 & \textbf{89.09} & 79.17 & 77.57 & 73.25 & 82.67 & 68.00 & 55.00 & 37.33 & 60.75\\
    & 4o-Mini & 67.50 & 67.00 & 59.00 & 85.71 & \textbf{84.57} & 84.37 & 72.04 & 84.33 & 70.00 & 59.00 & 44.67 & 64.50\\
    & 4o & 70.50 & 62.50 & 59.00 & 87.92 & 84.44 & 82.78 & 69.35 & 84.33 & 69.33 & 57.33 & 45.00 & 64.00\\
    \bottomrule
    \end{tabular}
    }
    \caption{Results of the evaluation of LLMs' instruction-following ability across different dimensions. `Overall' denotes the overall score. The best results in each dimension are highlighted in \textbf{bold}.}
    \label{tab:evaluation}
\end{table*}

\subsection{Instruction Generation Pipeline}
\label{sec:pipeline}

Building upon the multi-dimensional constraint framework, we introduce a controllable pipeline that transforms raw instructions into constrained versions that can be verified through code, as illustrated in Figure~\ref{fig:pipeline}.

\paragraph{Constraint Expansion}
Constraint expansion involves adding new constraints to a given instruction. Specifically, we randomly select a constraint category not yet covered and add one or two specific constraints from that category. This process progressively generates instructions across varying levels of constraint difficulty.

\paragraph{Conflict Detection}
Conflict detection ensures the soundness of instructions after constraint expansion. It consists of two checks: first, verifying that the newly specified constraints have been correctly incorporated; second, ensuring that the constraints do not conflict with each other (e.g., requiring a sentence to be entirely lowercase while also demanding the presence of uppercase words). If either check fails, the instruction is discarded. Otherwise, it is retained, and constraint expansion continues until difficulty Level IV is reached.

\paragraph{Instruction Rewriting}
Instruction rewriting enhances instruction diversity by transforming a given instruction to match a specified pattern. Specifically, we randomly select a constraint pattern and rewrite the instruction accordingly. When handling example-based constraints, we uniformly select three question-answer pairs that share the same constraint subcategories as the original instruction to serve as contextual examples.

\paragraph{Dataset}
We randomly sample data from ShareGPT\footnote{\url{https://www.sharegpt.com}} and generate 9,106 data instances using the aforementioned process. To ensure quality, each instance undergoes \emph{manual review},\footnote{More details can be found in Appendix~\ref{sec:manual}.} and Python validation code is written for every data point to enable automated evaluation. Statistical details of the data are presented in Table~\ref{tab:data}.\footnote{A comparison between MulDimIF and other approaches is provided in Appendix~\ref{sec:comparison}.}

\section{Evaluations of LLMs}

\subsection{Models}

Based on the test set described in Table~\ref{tab:data}, we conduct an evaluation of 18 LLMs from six model families, including four open-source and three closed-source. 
Among the open-source LLMs, we select \textit{LLaMA3.1-Instruct-8B} and \textit{LLaMA3.1-Instruct-70B} from the \textbf{LLaMA3.1 family}~\cite{LLaMA3.1};
\textit{DeepSeek-R1-Distill-LLaMA-8B} and \textit{DeepSeek-R1-Distill-LLaMA-70B} from the \textbf{DeepSeek-R1-Distill-LLaMA family}~\cite{DeepSeek-R1}; 
and \textit{Qwen3-8B}, \textit{Qwen3-14B}, and \textit{Qwen3-32B} from the \textbf{Qwen3 family}~\cite{Qwen3}.
Since Qwen3 includes specialized optimizations for reasoning, we evaluate its models in both non-reasoning mode (i.e., \textit{-Direct.}) and reasoning mode (i.e., \textit{-Reason.}).
For the closed-source LLMs, we include \textit{Gemini1.5-Flash} and \textit{Gemini1.5-Pro} from the \textbf{Gemini1.5 family}~\cite{Gemini1.5}; \textit{Claude3.5-Haiku} and \textit{Claude3.5-Sonnet} from the \textbf{Claude3.5 family}~\cite{Claude}; and \textit{GPT-3.5-Turbo}, \textit{GPT-4-Turbo}, \textit{GPT-4o-Mini}, and \textit{GPT-4o} from the \textbf{GPT family}~\cite{GPT-4}.\footnote{More information can be found in Appendix~\ref{sec:detail_model}.}

\subsection{Experimental Setup}
For instruction generation, we use GPT-4o.\footnote{The prompts are provided in Appendix~\ref{sec:prompt}.}
To ensure that each model’s capabilities are accurately represented, we use the built-in chat template for open-source models and the official API interface for closed-source models.
For consistency and reproducibility, we apply greedy decoding across all evaluations.
For LLMs that support a reasoning mode, we evaluate only the final answer and exclude the intermediate reasoning process.

\subsection{Main Results}

Table~\ref{tab:evaluation} summarizes the results of our multi-dimensional evaluation of various LLMs. Several key findings emerge from this analysis.

\textbf{There is substantial variation in current LLMs' ability to follow different constraint forms.} For constraint patterns, most models perform best on the example pattern, while performance steadily declines under the incorporation patter. This underscores the effectiveness of in-context learning~\cite{in-context} but also highlights the persistent difficulty of adhering to free-form constraints. Across constraint categories, models generally follow content-based constraints well but struggle with language- and length-related ones, often producing additional content to improve naturalness at the cost of constraint violations. In terms of difficulty, average accuracy drops sharply from 80.82\% at Level I to 36.76\% at Level IV, suggesting that current models face considerable challenges when handling multiple or more complex constraints.

\textbf{Scaling up models generally improves their ability to follow instructions, though exceptions are observed.} Within most families, larger models exhibit stronger adherence, especially in challenging scenarios such as level IV. This trend is consistent with prior findings on scaling laws for LLMs~\cite{scale_1, scale_2}. However, the Qwen3 and GPT families deviate from this pattern. For instance, Qwen3-32B-Direct. underperforms relative to Qwen3-14B-Direct., and GPT-4o falls short compared to GPT-4o-Mini. Such anomalies may reflect an alignment tax~\cite{instructgpt}, whereby optimization for broader capabilities comes at the expense of precision in instruction following.

\textbf{Optimizing reasoning capabilities has raised the lower bound of a model's instruction-following ability without enhancing its upper bound.} In the Qwen3 family, enhancements in reasoning mode primarily benefit previously weaker dimensions (e.g., incorporation, length, and Level IV) while providing little improvement in stronger ones (e.g., example, content, and Level I). Additionally, DeepSeek-R1-Distill-LLaMA-Instruct-8B shows reduced performance compared to LLaMA3.1-Instruct-8B. These findings underscore the urgent need for reasoning optimization strategies specifically designed to enhance instruction-following capabilities.

\begin{table}[!t]
\centering
\resizebox{0.85\linewidth}{!}
{
\begin{tabular}{lcc}
\toprule
\textbf{Dataset}  &  \textbf{Ability}   &  \textbf{\# Number} \\ \midrule
\rowcolor{gray!10} \textit{Training} & & \\
Ours & Instruction Following     &  7906 \\ \midrule
\rowcolor{gray!10} \textit{Test} & & \\
Ours & Instruction Following & 1200 \\
IFEval & Instruction Following & 541 \\
Multi-IF & Instruction Following & 13447 \\
MMLU & Knowledge & 14042 \\
GSM8K & Reasoning & 1319 \\
MATH & Reasoning & 5000 \\
HumanEval & Coding & 164 \\
MBPP & Coding & 257 \\
\bottomrule
\end{tabular}
}
\caption{Overview of the training and test datasets used when improving instruction-following capabilities.}
\vspace{-4mm}
\label{tab:data_grpo}
\end{table}

\section{Improvements of LLMs}
As shown in Table~\ref{tab:data}, we construct 7,906 single-turn, constraint-rich instructions designed for RL to enhance the instruction-following abilities of LLMs.\footnote{The full distribution of training data is provided in Table~\ref{tab:data}.} Our approach strengthens these abilities while preserving overall performance. Analysis indicates that the observed gains arise primarily from updates to attention modules, which increase the model's sensitivity to task-specific constraints.

\subsection{Test Set}
To compare model performance before and after RL, we evaluate four abilities:
1) \textbf{Instruction Following.} We evaluate on our test set, along with IFEval and Multi-IF.
2) \textbf{Knowledge.} General knowledge is evaluated using MMLU~\cite{MMLU}.
3) \textbf{Reasoning.} We use GSM8K~\cite{GSM8K} and MATH~\cite{MATH} to measure logical and mathematical reasoning.
4) \textbf{Coding.} Programming ability is evaluated with HumanEval~\cite{HumanEval} and MBPP~\cite{MBPP}.
The information of the data is shown in Table~\ref{tab:data_grpo}.

\begin{figure*}[!t]
    \centering
    \includegraphics[width=\linewidth]{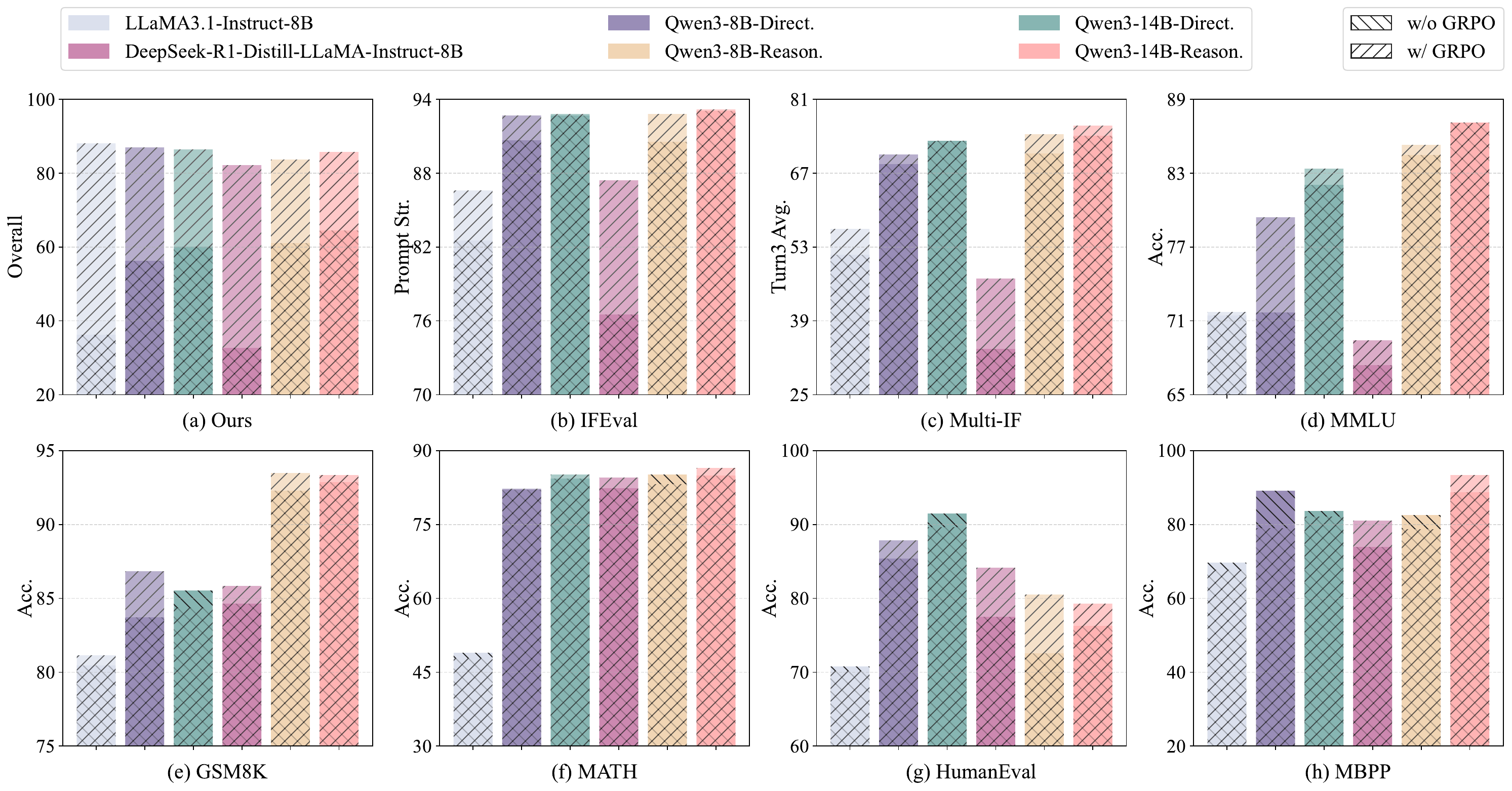}
    \caption{Performance comparison of each LLM on the test sets before and after applying GRPO.}
    \label{fig:improve}
\end{figure*}

\subsection{Experimental Setup}

We conduct experiments on six LLMs without more than 14 billion parameters, applying the GRPO algorithm. The setup includes a batch size of 1,024, mini-batch size of 512, 32 rollouts per update, and a learning rate of 1e-6. We use a sampling temperature of 0.7 and set the maximum output length to 8,192 tokens. The reward function is defined as the number of constraints satisfied in the output. Training is performed for one epoch on eight NVIDIA A800 GPUs. For evaluation, we apply each model's official chat template and use greedy decoding for consistency.

\subsection{Results and Analysis}


\paragraph{Performance Improvements}
Figure~\ref{fig:improve} presents the performance of individual LLMs before and after applying GRPO.\footnote{Detailed results are provided in Appendix~\ref{sec:detail_result}.} The results demonstrate substantial performance gains on our custom test set, with LLaMA3.1-Instruct-8B notably outperforming other models. Importantly, these improvements extend to out-of-domain instruction-following benchmarks and significantly enhance performance in multi-turn dialogue scenarios (i.e., Multi-IF), despite training being conducted solely on single-turn data. This suggests that the data generated by our multi-dimensional constraint framework exhibits strong generalization ability. Furthermore, although our training is focused on improving instruction-following abilities, it does not degrade general-purpose performance. On general benchmarks, post-GRPO LLMs maintain parity with their original counterparts and, in some cases (e.g., MMLU), show clear improvements. These findings indicate that our pipeline produces data that is both compatible with and complementary to existing training corpora, enabling straightforward performance enhancements when integrated into current LLMs.

\begin{figure}[!t]
    \centering
    \begin{subfigure}[!t]{0.45\linewidth}
        \centering
        \includegraphics[width=\linewidth]{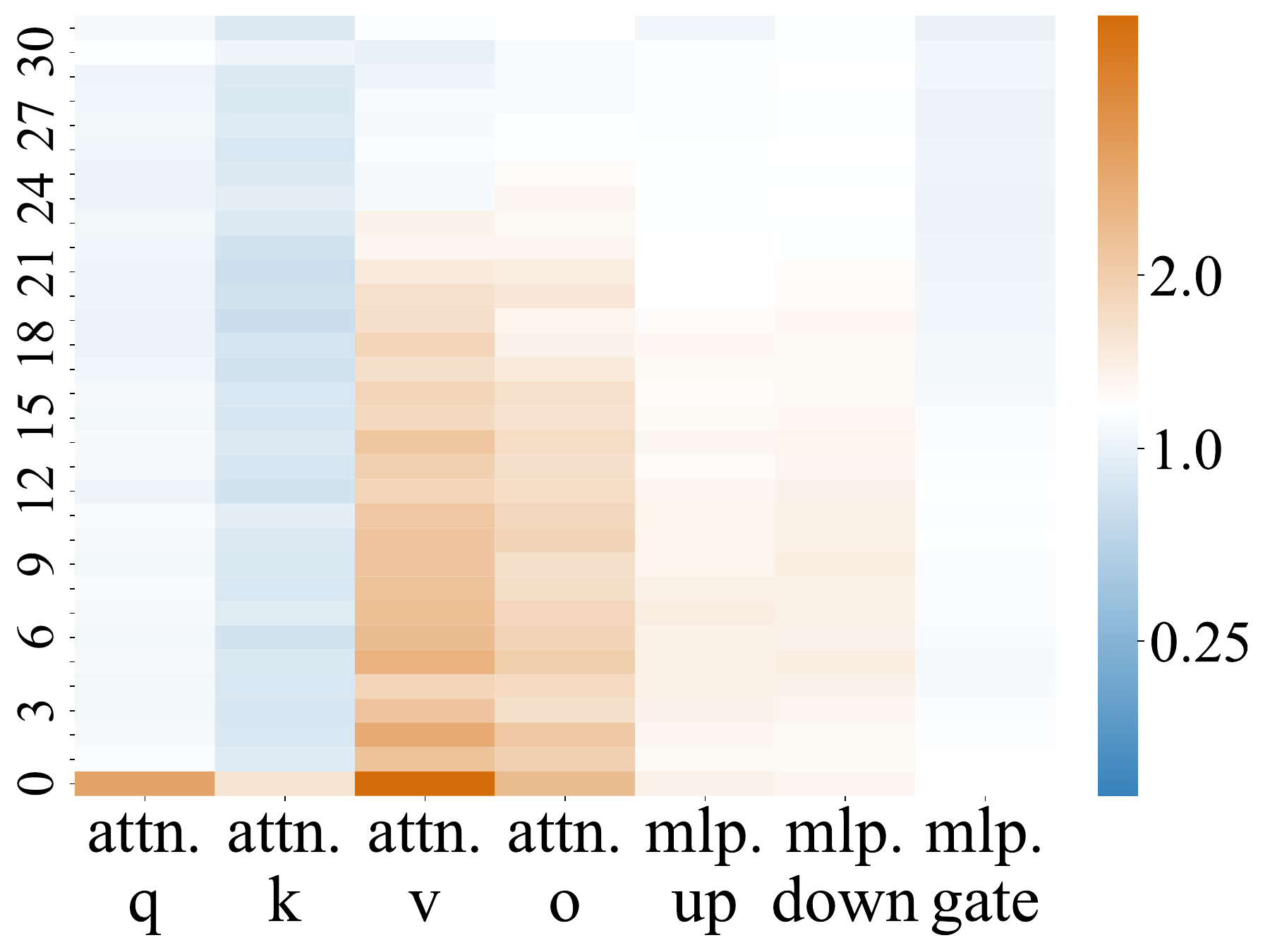}
        \caption{LLaMA3.1-Instruct\\-8B}
        \label{fig:param-llama3.1-8b}
    \end{subfigure}
    \quad
    \begin{subfigure}[!t]{0.45\linewidth}
        \centering
        \includegraphics[width=\linewidth]{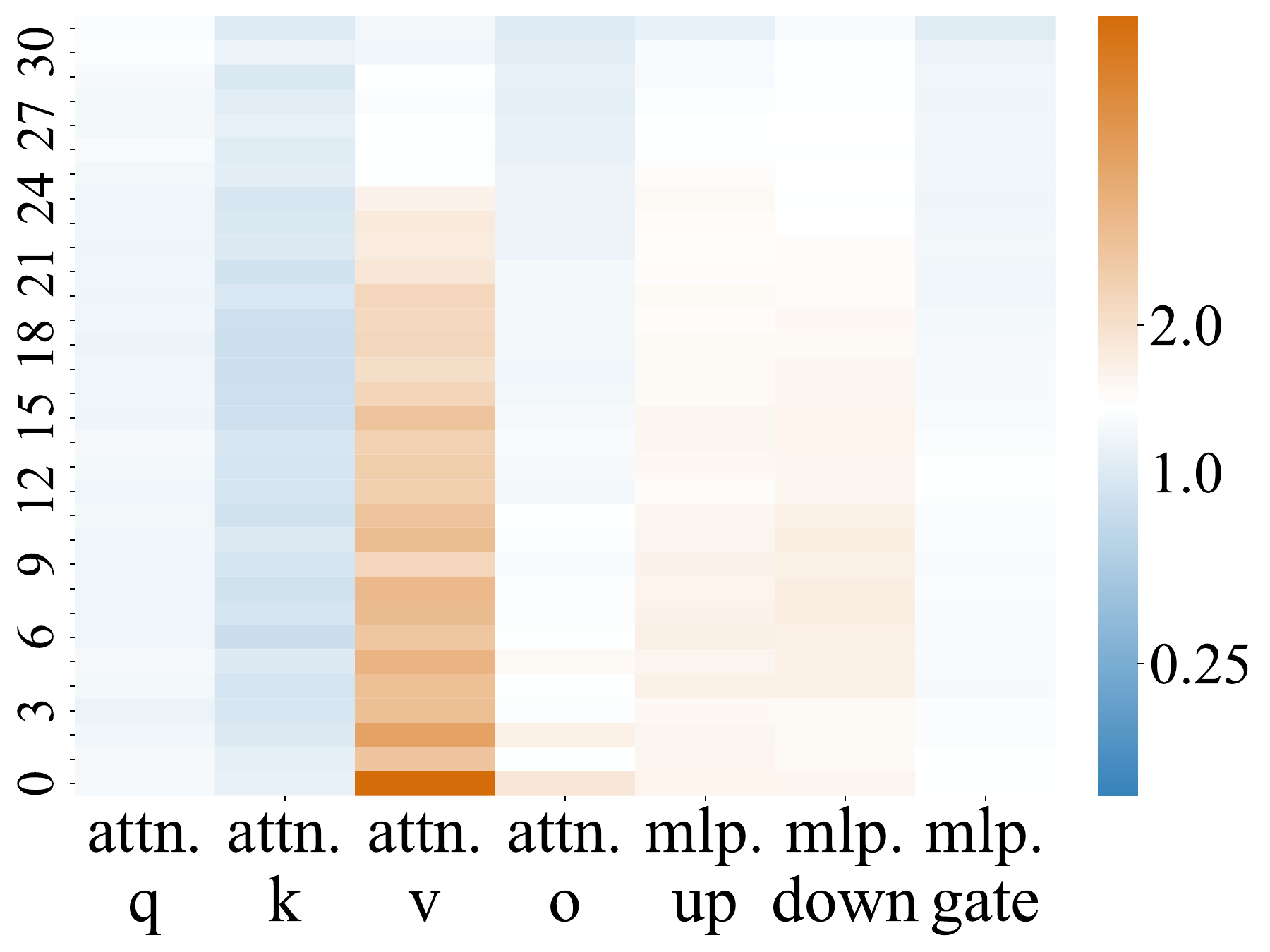}
        \caption{\centering DeepSeek-R1-Distill\\-LLaMA-Instruct-8B}
        \label{fig:param-deepseek-llama-8b}
    \end{subfigure}
    \quad
    \begin{subfigure}[!t]{0.45\linewidth}
        \centering
        \includegraphics[width=\linewidth]{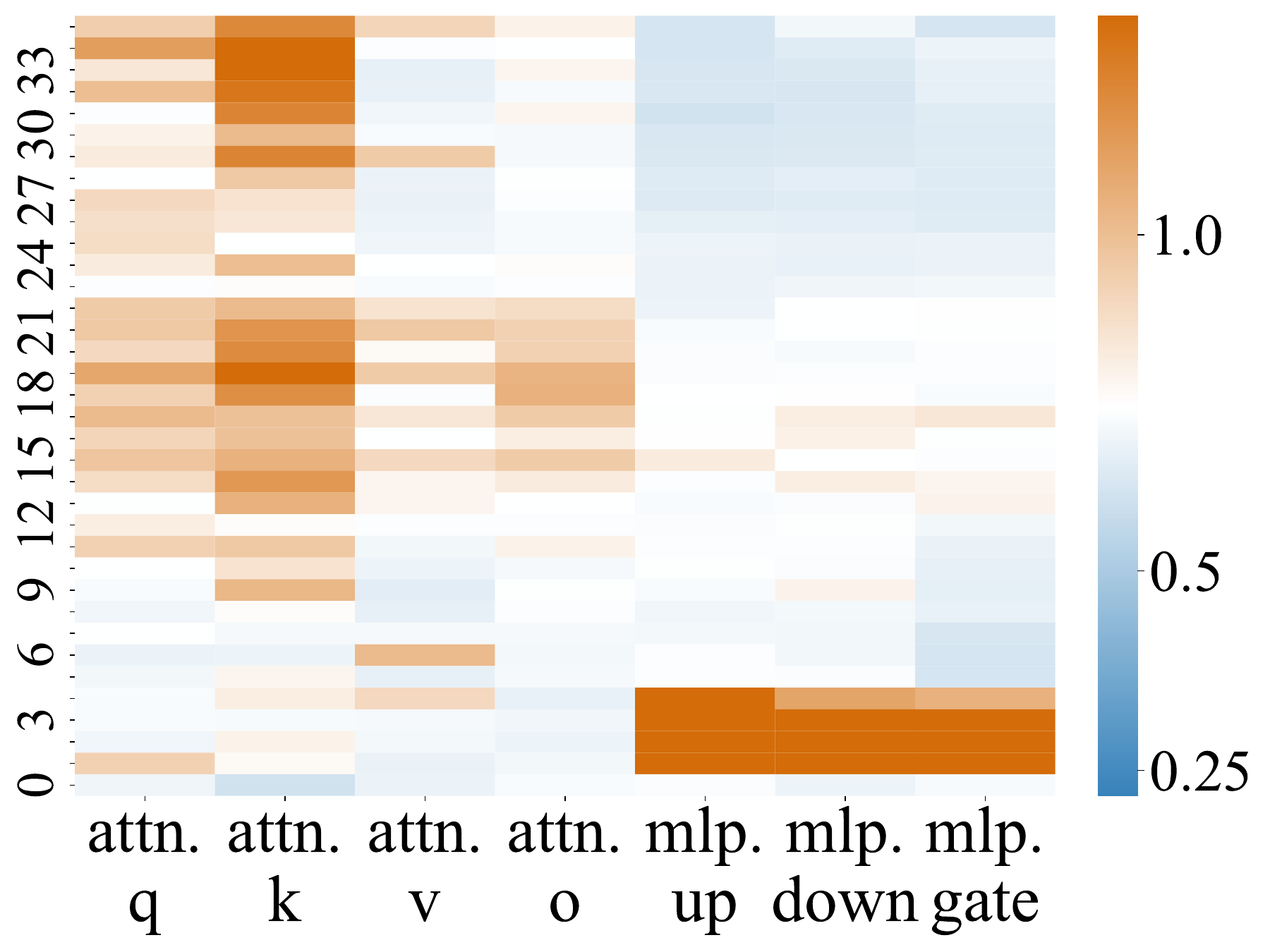}
        \caption{Qwen3-8B-Direct.}
        \label{fig:param-qwen2.5-7b}
    \end{subfigure}
    \quad
    \begin{subfigure}[!t]{0.45\linewidth}
        \centering
        \includegraphics[width=\linewidth]{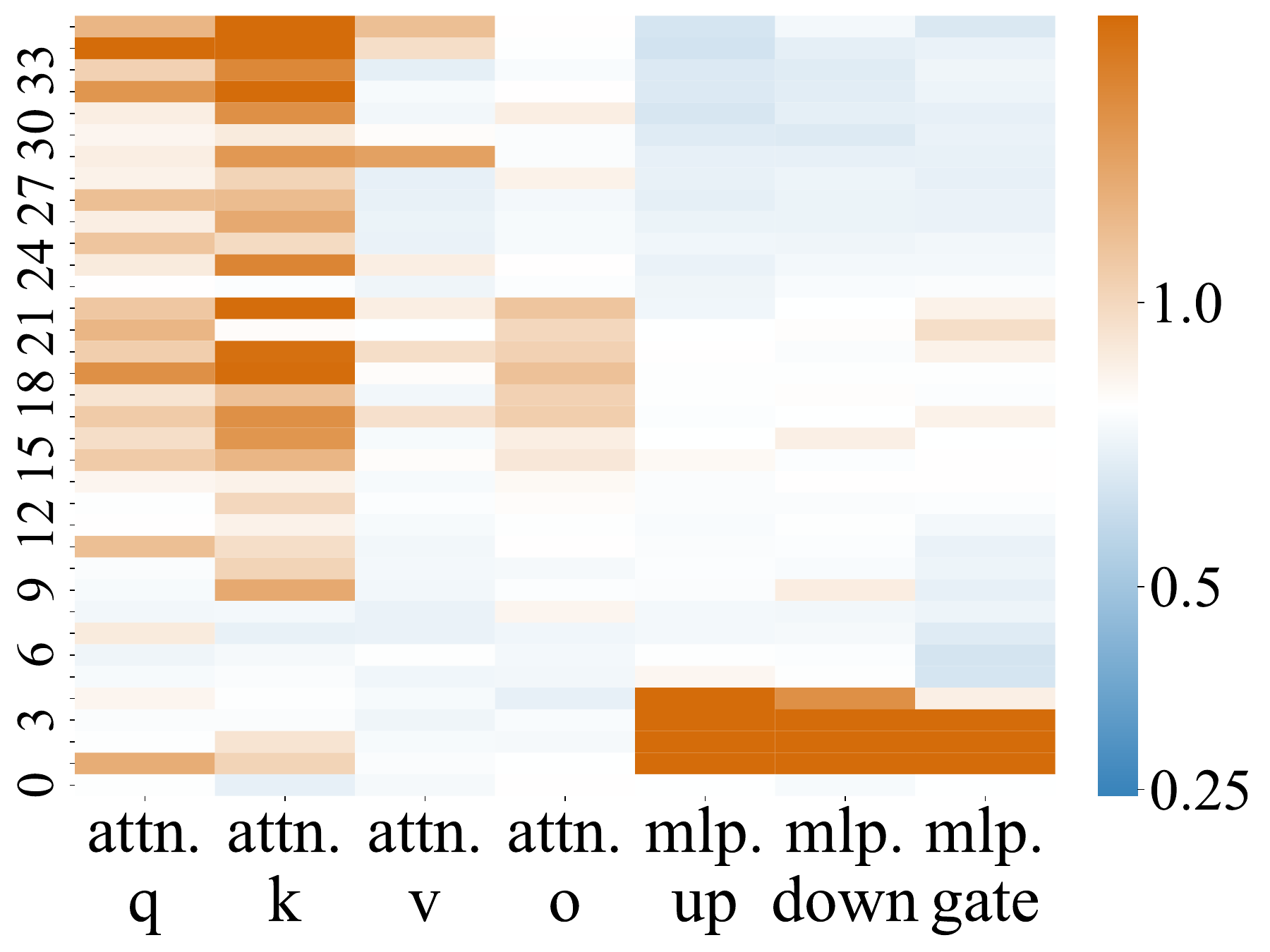}
        \caption{Qwen3-8B-Reason.}
        \label{fig:param-deepseek-qwen-7b}
    \end{subfigure}
    \quad
    \begin{subfigure}[!t]{0.45\linewidth}
        \centering
        \includegraphics[width=\linewidth]{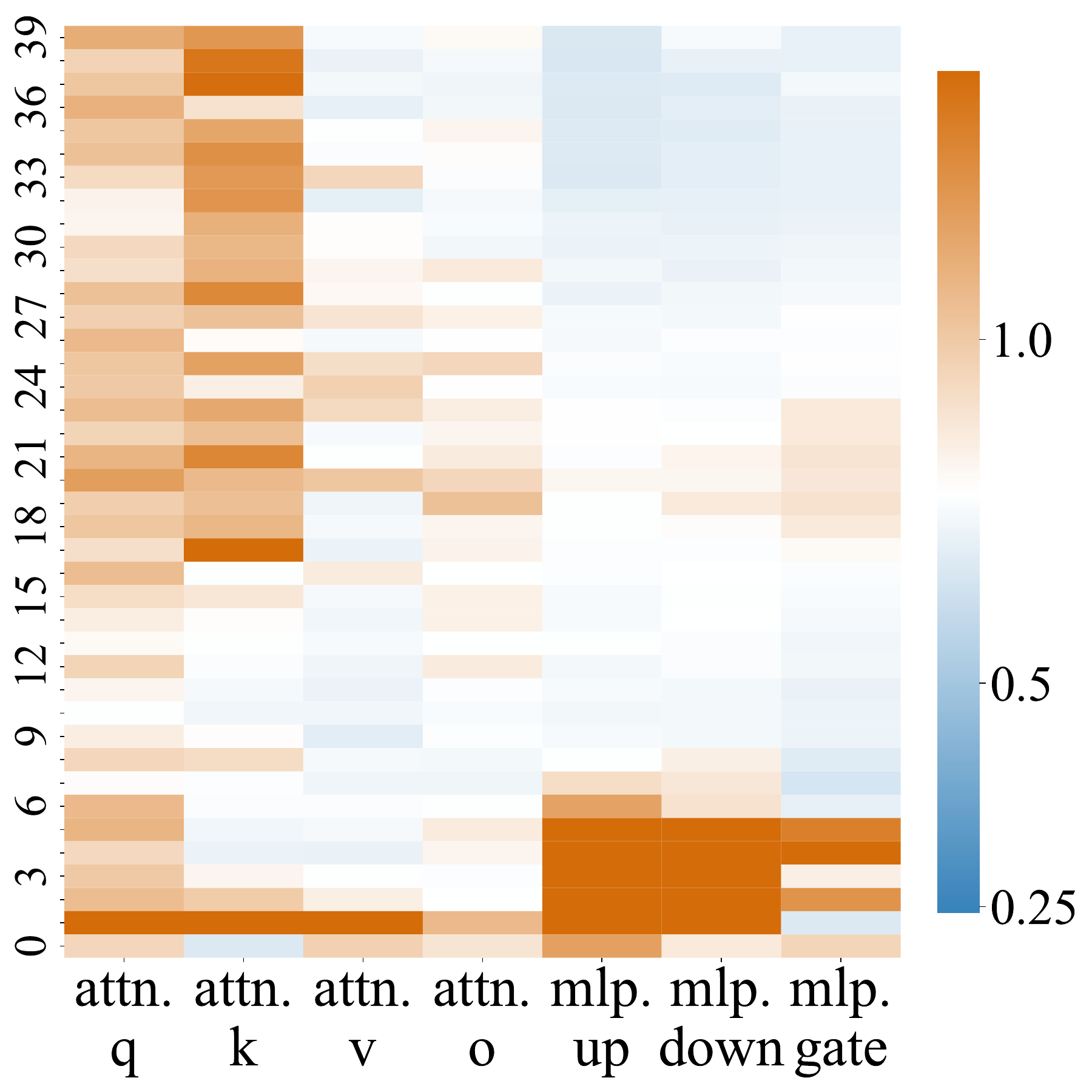}
        \caption{Qwen3-14B-Direct.}
        \label{fig:param-qwen2.5-14b}
    \end{subfigure}
    \quad
    \begin{subfigure}[!t]{0.45\linewidth}
        \centering
        \includegraphics[width=\linewidth]{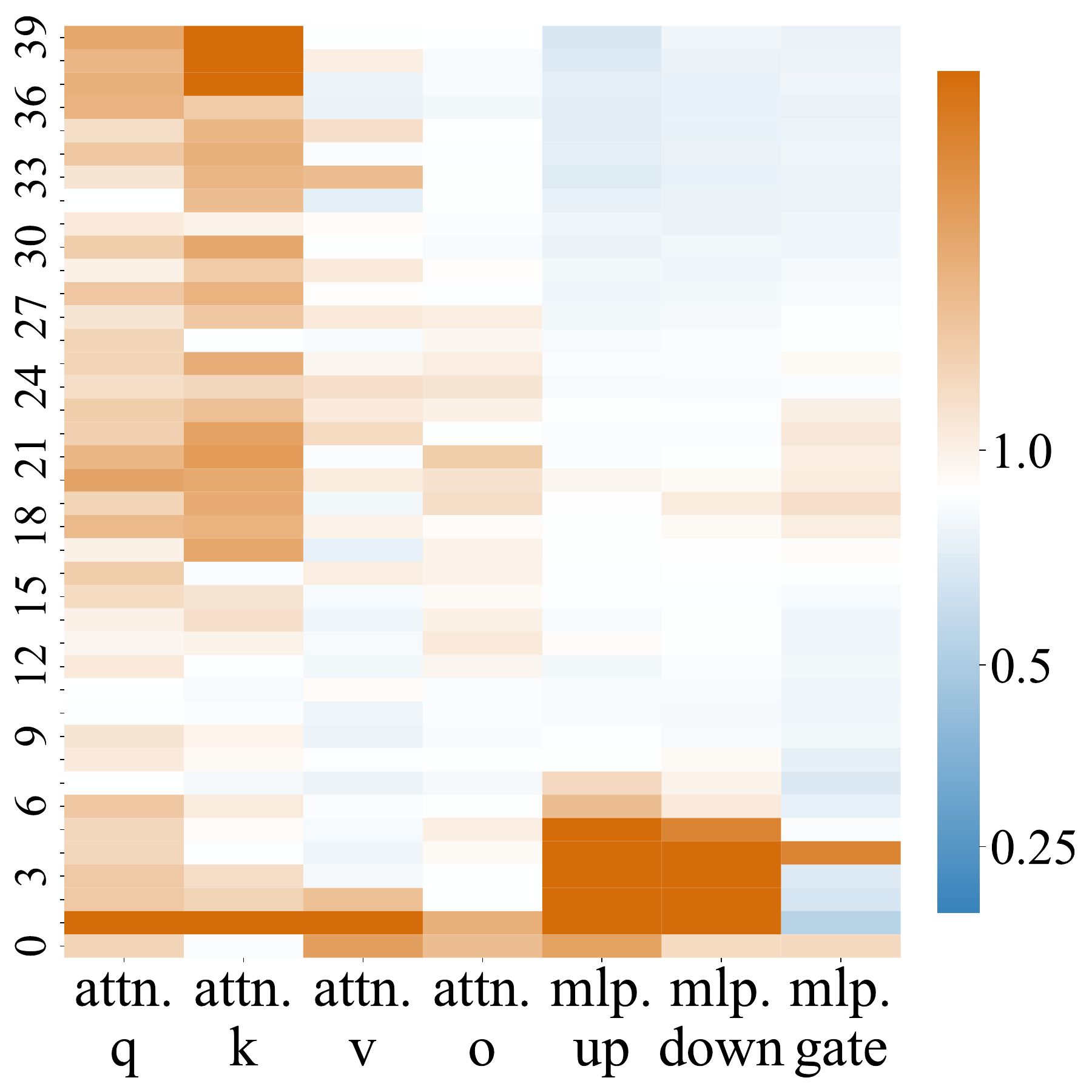}
        \caption{Qwen3-14B-Reason.}
        \label{fig:param-deepseek-qwen-14b}
    \end{subfigure}
    
    \caption{Parameter change rates of LLMs after GRPO relative to the original ones across different modules.
    }
    \label{fig:parameter}
\end{figure}

\paragraph{Parameter-Level Analysis}
To better understand the sources of performance improvement, we conduct a parameter-level analysis. We compute the relative change rates of model parameters following GRPO, broken down by model modules, and summarize the results in Figure~\ref{fig:parameter}. Notably, the most substantial updates occur in attention modules, suggesting that GRPO primarily refines the model's attention mechanisms. These changes are distributed relatively uniformly across layers, indicating a global rather than localized adjustment. Overall, applying GRPO with our data effectively tunes the model's attention distribution, enhancing its ability to identify and focus on critical input information and thereby improving its instruction-following and general performance.

\begin{table*}[t!]
\centering
\resizebox{\linewidth}{!}
{
\begin{tabular}{p{0.5\linewidth}<{\centering}p{0.5\linewidth}<{\centering}}
\toprule
\textbf{Case (w/o GRPO)} & \textbf{Case (w/ GRPO)}  \\ \midrule
\rowcolor{gray!10} \multicolumn{2}{p{\linewidth}}{\textit{\textbf{LLaMA3.1-Instruct-8B:} After applying GRPO, the model becomes more attentive to nuanced constraints, especially specific details like the `5 words' requirement, ensuring it meets the length constraint.}} \\
\begin{minipage}[h]{\linewidth}
		\centering
		{\includegraphics[width=\linewidth]{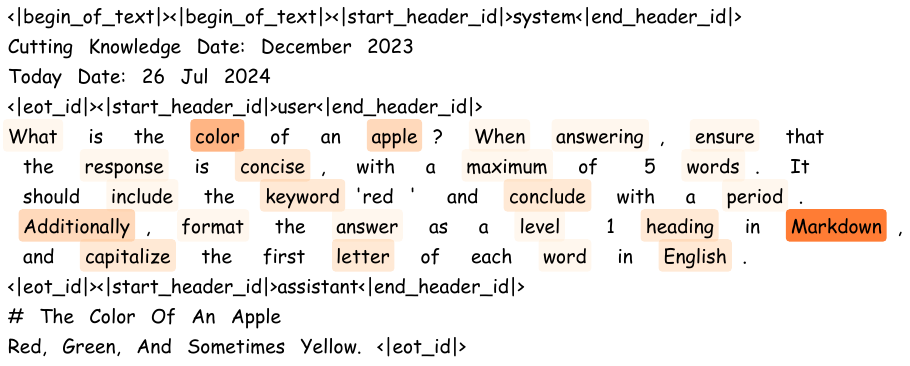}}
\end{minipage}
 & 
 \begin{minipage}[h]{\linewidth}
		\centering
		{\includegraphics[width=\linewidth]{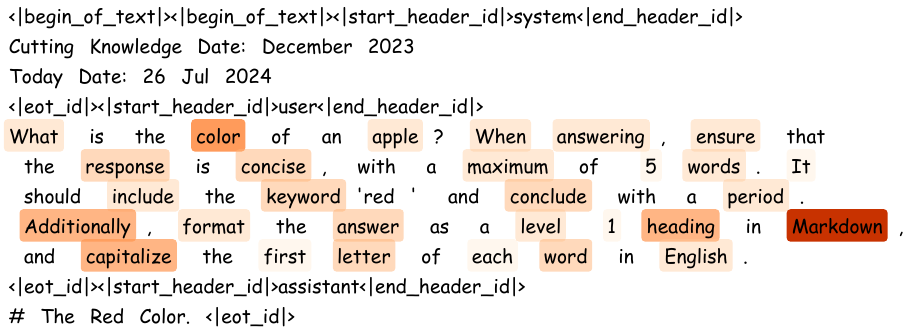}}
\end{minipage}
\\ 
\midrule
\rowcolor{gray!10} \multicolumn{2}{p{\linewidth}}{\textit{\textbf{Qwen3-8B-Direct.:} After applying GRPO, the model increased its attention to details such as `capitalized' and `each word', ensuring that every word in the output is capitalized.}} \\
\begin{minipage}[h]{\linewidth}
		\centering
		{\includegraphics[width=\linewidth]{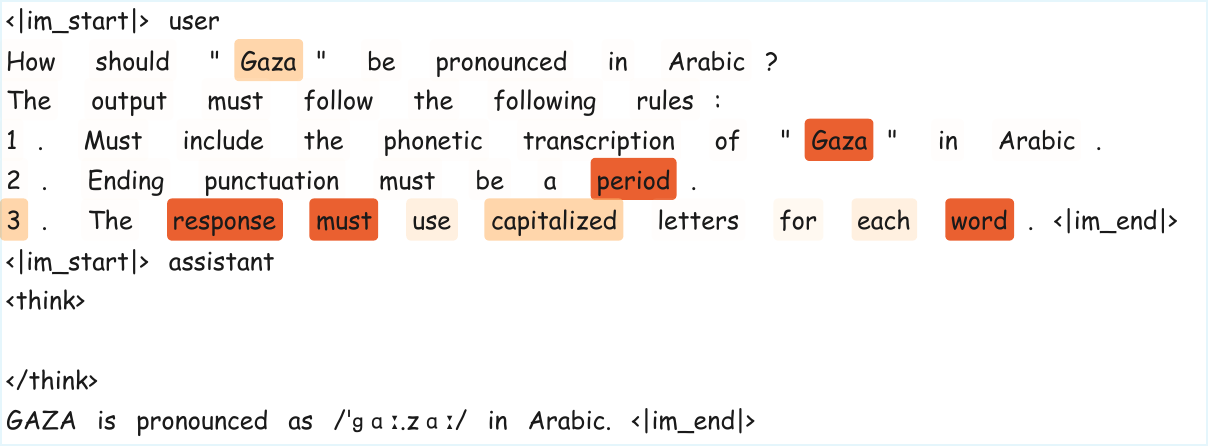}}
\end{minipage}
 & 
 \begin{minipage}[h]{\linewidth}
		\centering
		{\includegraphics[width=\linewidth]{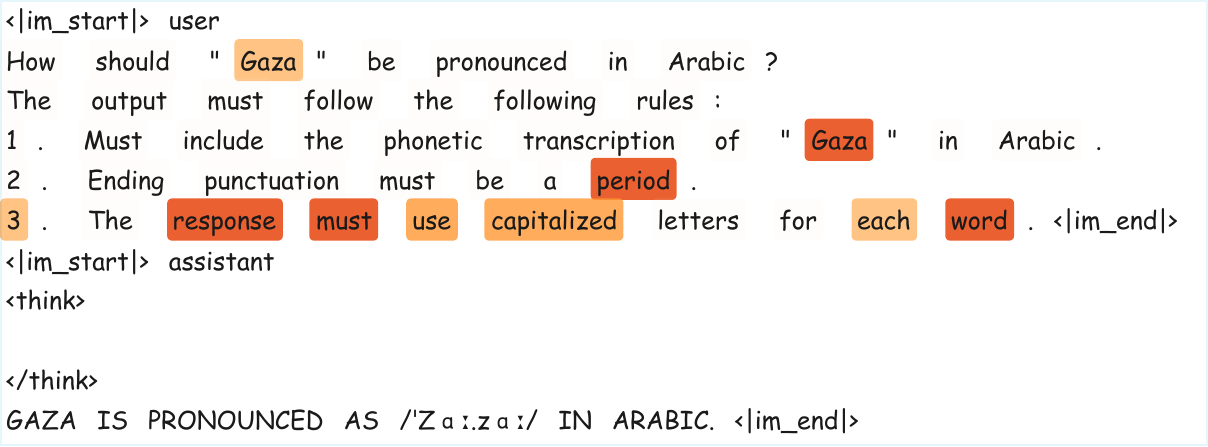}}
\end{minipage}
\\ 
\midrule
\rowcolor{gray!10} \multicolumn{2}{p{\linewidth}}{\textit{\textbf{DeepSeek-R1-Distill-LLaMA-Instruct-8B:} After applying GRPO, the model places greater emphasis on factors such as length, keywords, and content, enabling it to meet multiple complex constraints.}} \\
\begin{minipage}[h]{\linewidth}
		\centering
		{\includegraphics[width=\linewidth]{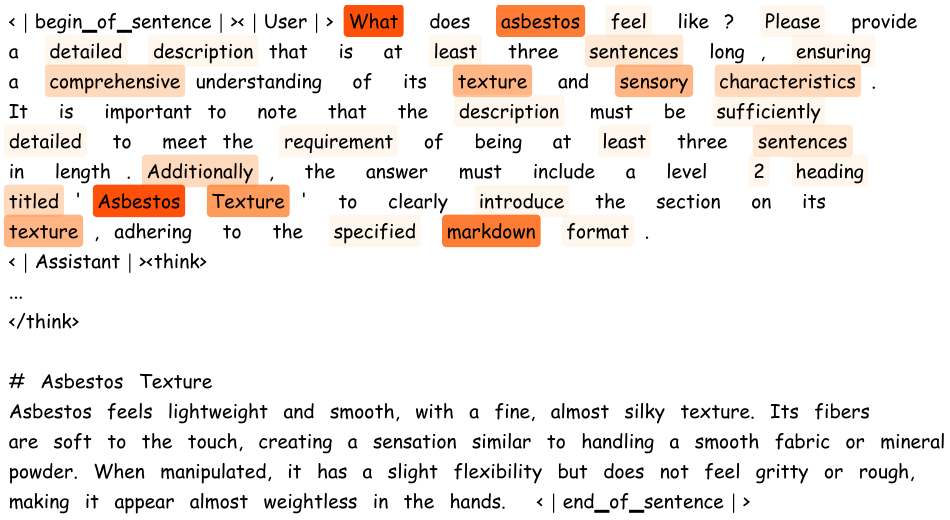}}
\end{minipage}
 & 
 \begin{minipage}[h]{\linewidth}
		\centering
		{\includegraphics[width=\linewidth]{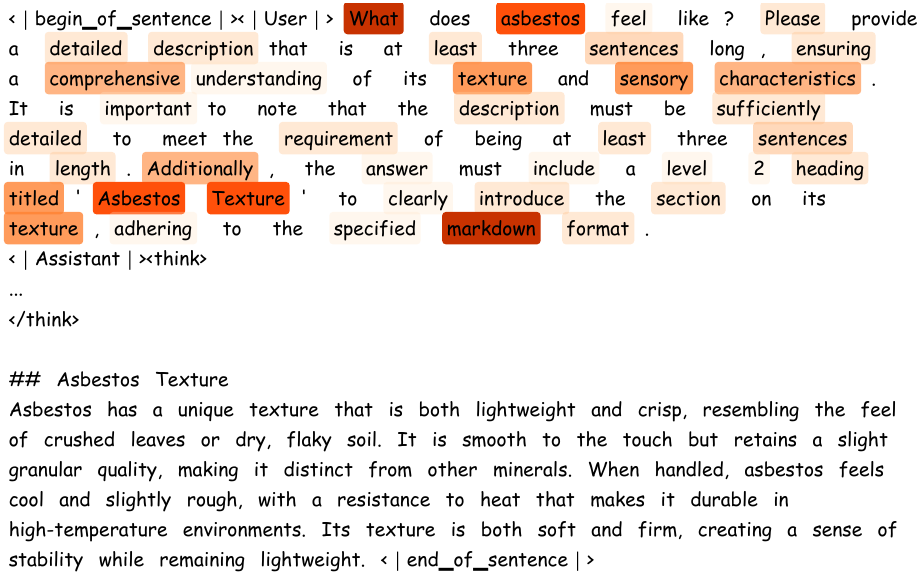}}
\end{minipage}
\\ 
\midrule
\rowcolor{gray!10} \multicolumn{2}{p{\linewidth}}{\textit{\textbf{Qwen3-8B-Reason.:} After applying GRPO, the model places stronger emphasis on word count and sentence length, ensuring compliance with length constraints.}} \\
\begin{minipage}[h]{\linewidth}
		\centering
		{\includegraphics[width=\linewidth]{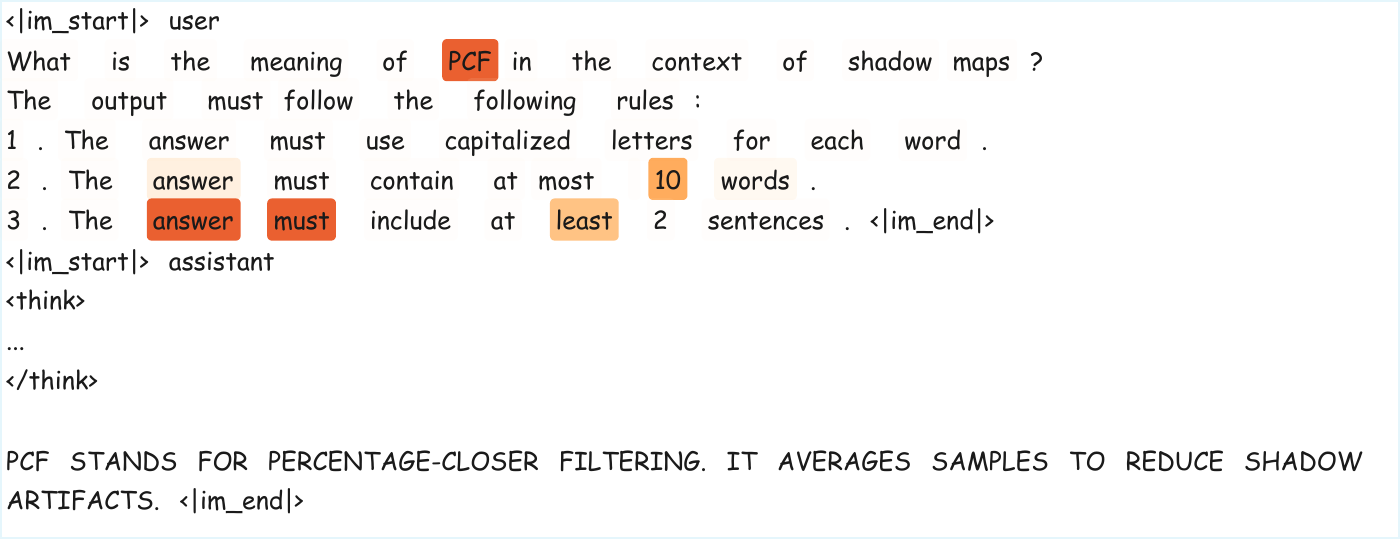}}
\end{minipage}
 & 
 \begin{minipage}[h]{\linewidth}
		\centering
		{\includegraphics[width=\linewidth]{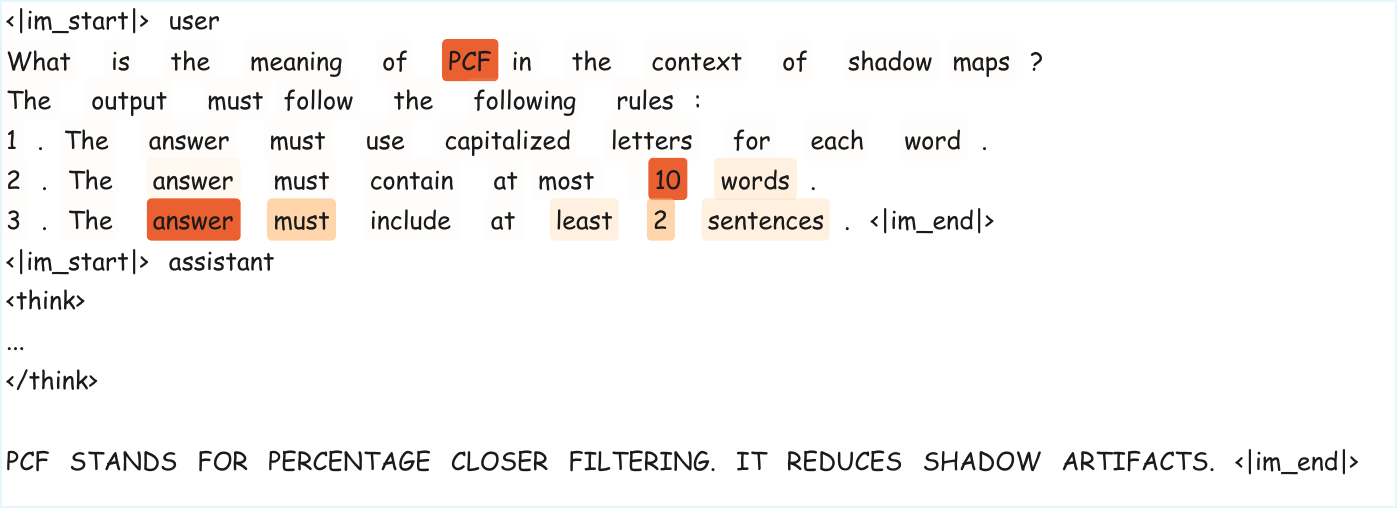}}
\end{minipage}
\\ 
\bottomrule
\end{tabular}
}
\caption{Visualization of the importance of each input token to the output. Darker colors indicate greater significance.}
\label{tab:case}
\vspace{-4mm}
\end{table*}

\paragraph{Case Studies}
To visualize how modifications in attention mechanisms affect model behavior, we adopt the information flow analysis method proposed by~\citet{information-flow}. We compute the importance of each input token with respect to the model's output and present representative visualizations in Table~\ref{tab:case}. After applying GRPO, the importance of constraint-related tokens increases, while the influence of irrelevant tokens diminishes; the relevance of core problem components remains largely unchanged. This suggests that the model has improved its ability to identify and prioritize constraint-related information without sacrificing its understanding of the overall input. Additionally, the reduced attention to irrelevant tokens may help minimize distraction from non-essential elements, which could explain the observed stability or gains in general task performance. These case studies further validate the effectiveness of our proposed framework and the utility of the data it produces.

\section{Conclusion}
In this paper, we propose a multi-dimensional constraint framework that categorizes instructions based on constraint pattern, constraint category, and constraint difficulty. Building on it, we design a controllable instruction generation pipeline that transforms raw instructions into constraint-based variants through the processes of constraint expansion, conflict detection, and instruction rewriting. Using this pipeline, we generate 9,106 instances. We then conduct a comprehensive evaluation of 18 LLMs from six model families and apply the GRPO algorithm to enhance LLMs' abilities. The results show that models trained with our data achieve significant improvements while preserving general capabilities. Parameter-level analysis and case studies suggest that these gains primarily arise from increased model sensitivity to constraint-relevant information.

\section*{Limitations}
Although we evaluate and enhance the instruction-following ability of current LLMs using constructive data and analyze the underlying reasons for their improvement, our work still has two key limitations.
On one hand, due to resource constraints, we assess model performance before and after RL only on a limited set of representative datasets, which may not fully capture changes in the models’ original abilities. Nevertheless, our results show that performance on these widely used benchmarks remains largely stable, underscoring the robustness of our constructed data.
On the other hand, since our focus is primarily on enhancing instruction-following abilities, we do not explore the effects of applying our method to domain-specific datasets. However, because our approach can convert any instruction into a constraint-based version, and case studies confirm that the model retains its focus on the core problem components, we believe that applying this method to other domains (e.g., reasoning, coding) can also yield additional performance gains.

\section*{Acknowledgments}
The authors wish to thank the anonymous reviewers for their helpful comments. This work was partially funded by National Natural Science Foundation of China (No. 62476061, 62576106, 62376061).

\bibliography{custom}



\appendix

\section{Comparison of MulDimIF with Existing Approaches}
\label{sec:comparison}

\begin{table*}[!t]
\centering
\resizebox{\linewidth}{!}
{
\begin{tabular}{lccccccccc}
\toprule
\multirow{2}*{\textbf{Method}} & \textbf{\# Train} & \textbf{\# Test} & \textbf{Turn} & \textbf{Constraint} & \textbf{Automated} & \textbf{Code} & \textbf{Incorporation} & \textbf{Listing} & \textbf{Example} \\
& \textbf{Data} & \textbf{Data} & \textbf{Category} & \textbf{Category} & \textbf{Pipeline} & \textbf{Verifiable} & \textbf{Pattern} & \textbf{Pattern} & \textbf{Pattern} \\
\midrule
IFEval~\cite{IFEval} & 0 & 541 & Single & 9 & {\color{GREEN}$\checkmark$} & {\color{GREEN}$\checkmark$} & {\color{GREEN}$\checkmark$} & {\color{red}$\times$} & {\color{red}$\times$} \\
Conifer~\cite{Conifer} & 13,606 & 0 & Single & -- & {\color{GREEN}$\checkmark$} & {\color{red}$\times$} & {\color{GREEN}$\checkmark$} & {\color{red}$\times$} & {\color{red}$\times$} \\
CI-DPO~\cite{CI-DPO} & 2,781 & 0 & Single & -- & {\color{red}$\times$} & {\color{GREEN}$\checkmark$} & {\color{GREEN}$\checkmark$} & {\color{red}$\times$} & {\color{red}$\times$} \\
IOPO~\cite{IOPO} & 119,345 & 1,042 & Single & 4 & {\color{GREEN}$\checkmark$} & {\color{red}$\times$} & {\color{GREEN}$\checkmark$} & {\color{red}$\times$} & {\color{red}$\times$} \\
FollowBench~\cite{FollowBench} & 0 & 820 & Single & 4 & {\color{red}$\times$} & {\color{red}$\times$} & {\color{GREEN}$\checkmark$} & {\color{red}$\times$} & {\color{GREEN}$\checkmark$} \\
Multi-IF~\cite{Multi-IF} & 0 & 13,447 & Multi & 9 & {\color{GREEN}$\checkmark$} & {\color{GREEN}$\checkmark$} & {\color{GREEN}$\checkmark$} & {\color{red}$\times$} & {\color{red}$\times$} \\
AutoIF~\cite{AutoIF} & 61,492 & 0 & Single & -- & {\color{GREEN}$\checkmark$} & {\color{GREEN}$\checkmark$} & {\color{GREEN}$\checkmark$} & {\color{red}$\times$} & {\color{red}$\times$} \\
AGENTIF~\cite{AGENTIF} & 0 & 707 & Multi & 3 & {\color{red}$\times$} & {\color{red}$\times$} & {\color{GREEN}$\checkmark$} & {\color{red}$\times$} & {\color{GREEN}$\checkmark$} \\
\rowcolor{gray!10} MulDimIF (Ours) & 7,906 & 1,200 & Single & 4 & {\color{GREEN}$\checkmark$} & {\color{GREEN}$\checkmark$} & {\color{GREEN}$\checkmark$} & {\color{GREEN}$\checkmark$} & {\color{GREEN}$\checkmark$} \\
\bottomrule
\end{tabular}
}
\caption{Comparison of MulDimIF with existing approaches.}
\label{tab:comparison}
\end{table*}

To visually highlight the similarities and differences between our approach and existing methods, we provide a comprehensive comparison in Table~\ref{tab:comparison}. This comparison shows that MulDimIF offers several key advantages:
\begin{itemize}
    \item \textbf{Unified generation of both training and testing data:} Whereas most prior methods focus on only one, our framework systematically constructs both in a principled and consistent way.
    \item \textbf{Broad coverage of constraint categories and patterns:} MulDimIF supports a richer and more comprehensive set of constraint dimensions.
    \item \textbf{Fully automated data construction:} This enables scalable data augmentation while minimizing manual effort.
    \item \textbf{Code-verifiable outputs:} The generated data can be validated automatically, strengthening the rigor and reliability of evaluation.
\end{itemize}

\section{Details about Manual Review}
\label{sec:manual}

To ensure the quality of the constructed data, we carry out a thorough manual inspection process as outlined below. Six trained annotators (computer science graduate and undergraduate students) spend four weeks verifying and refining all 9,106 data instances along with their corresponding validation code.

Annotators first examine each generated instance to confirm that all specified constraints are correctly represented, mutually compatible, and expressed unambiguously. When issues are identified, the instance is revised and rechecked. Each item undergoes cross-validation by at least two annotators and is retained only when both agree that it is error-free.

After the data are finalized, we group the instances by constraint category. DeepSeek-V3~\cite{DeepSeek-V3} then generates initial validation code for each group. Annotators review and refine this code to ensure logical correctness and full coverage of the data. A second annotator independently evaluates the revised code. Code is accepted only when both reviewers approve it; otherwise, it is returned for further revision.

These procedures are designed to maximize the quality of both the dataset and the validation code.


\section{Examples for Different Forms of Constraints}
\label{sec:examples}

To illustrate the instructions generated by our method, we present a selection in Table~\ref{tab:examples}, annotated with their corresponding dimension information. These examples demonstrate that our approach overcomes the templating limitations of prior methods and better captures the diversity of real-world needs. Additionally, the multi-dimensional categorization enables a more fine-grained performance analysis.

\section{Detailed Information of Models}
\label{sec:detail_model}

We conduct an evaluation of 18 LLMs from seven families, including four open-source and three closed-source, that collectively represent the capabilities of current LLMs.

\paragraph{Open-Source LLMs}  
We evaluate eleven open-source LLMs across four model families:

\begin{itemize}
    \item \textbf{LLaMA3.1 family}, developed by Meta, comprises open-source models that demonstrate strong performance in general knowledge and multilingual translation. We evaluate \textit{LLaMA3.1-Instruct-8B} and \textit{LLaMA3.1-Instruct-70B}.

    \item \textbf{DeepSeek-R1-Distill-LLaMA family} consists of models distilled from the LLaMA family using data generated by DeepSeek-R1~\cite{DeepSeek-R1}. These models exhibit notable gains in mathematical performance. We evaluate \textit{DeepSeek-R1-Distill-LLaMA-8B} and \textit{DeepSeek-R1-Distill-LLaMA-70B}.

    \item \textbf{Qwen3 family}, released by Alibaba, is the latest version of the Qwen model family. It includes both dense and Mixture-of-Experts (MoE)~\cite{MoE} architectures, with parameter sizes ranging from 0.6 to 235 billion. In our study, we evaluate \textit{Qwen3-8B}, \textit{Qwen3-14B}, and \textit{Qwen3-32B}. Since Qwen3 includes specialized optimizations for reasoning, we evaluate its models in both non-reasoning mode (i.e., \textit{-Direct.}) and reasoning mode (i.e., \textit{-Reason.}).
\end{itemize}

\paragraph{Closed-Source LLMs}  
We evaluate eight closed-source LLMs across three model families:

\begin{itemize}
    \item \textbf{Gemini1.5 family}, developed by Google, represents a new generation of models built on the MoE architecture. These models demonstrate enhanced performance, particularly in long-context understanding across multiple modalities. We evaluate \textit{Gemini1.5-Flash} and \textit{Gemini1.5-Pro}.

    \item \textbf{Claude3.5 family}, developed by Anthropic, features models with strong general-purpose capabilities and coding capabilities. We evaluate \textit{Claude3.5-Haiku} and \textit{Claude3.5-Sonnet}.

    \item \textbf{GPT family}, released by OpenAI, includes models that represent the current frontier in LLM development. We evaluate \textit{GPT-3.5-Turbo}, \textit{GPT-4-Turbo}, \textit{GPT-4o-Mini}, and \textit{GPT-4o}.
\end{itemize}

\section{Detailed Results}
\label{sec:detail_result}

\subsection{Details of Training Process}
\label{sec:detail_train}

Table~\ref{tab:train-process} and Table~\ref{tab:test-process} present the model’s critic scores and its performance on MulDimIF, respectively, as training progresses. The results show a clear pattern of gradual convergence over time. At the same time, the consistently upward trend further demonstrates steady performance improvements throughout training, underscoring the effectiveness of our data.


\subsection{Details of Final Results}
\label{sec:detail_final_result}

Table~\ref{tab:improve_result_ours} to Table~\ref{tab:opencampass} present the detailed performance of the LLMs on each test set before and after GRPO. In these tables, $\Delta$ denotes the performance difference between the post-GRPO and pre-GRPO models, with positive values highlighted in \textcolor{GREEN}{green} and negative values in \textcolor{RED}{red}. The results demonstrate that applying our data for GRPO substantially enhances the models' capabilities across all aspects, highlighting the effectiveness of our data.

\section{Prompts for Instruction Generation}
\label{sec:prompt}

Our automated instruction generation pipeline includes constraint expansion, conflict detection, and instruction rewriting, with several steps utilizing GPT-4o. The corresponding prompts are listed from Figure~\ref{fig:prompt-ce} to Figure~\ref{fig:prompt-iri}.
Some of the information used in the prompts is presented in Table~\ref{tab:constraints} and Table~\ref{tab:cc_List}.

\begin{table*}[h]
\centering
\resizebox{\linewidth}{!}
{
\begin{tabular}{m{0.7\linewidth}cm{0.1\linewidth}<{\centering}c}
\toprule
\multicolumn{1}{c}{\textbf{Instruction}} & \textbf{Pattern} & {\textbf{Category}} & \textbf{Difficulty}\\

\midrule
\# Example 1:
\newline **Question**: What Is The Room? The answer must use capitalized letters for each word.
\newline **Answer**: The Room Is A Living Room.
\newline 
\newline \# Example 2:
\newline **Question**: Describe life in 1980. The answer must use capitalized letters for each word.
\newline **Answer**: Life In 1980 Was Characterized By The Rise Of Personal Computers, Vibrant Music Scenes With Genres Like Disco And Rock, And Fashion Trends Featuring Bold Colors And Styles. People \newline Enjoyed Watching Movies On VHS Tapes And Listening To Music On Cassette Players.
\newline 
\newline \# Example 3:
\newline **Question**: Can I insert PDFs or documents? The answer must use capitalized letters for each word.
\newline **Answer**: Yes, You Can Insert PDFs Or Documents.
\newline 
\newline **Question**: Write a script about Matt Rhule's 0-3 record so far. The answer must use capitalized letters for each word.
 & Example 
 & Language
 & Level I \\

\midrule
\# Example 1:
\newline **Question**: What are the best books to learn Cybersecurity in 2023? The answer must use capitalized letters for each word and must include the keyword `Cybersecurity'.
\newline **Answer**: ``Cybersecurity For Beginners'' And ``The Art Of Invisibility'' Are Among The Best Books To Learn Cybersecurity In 2023.
\newline 
\newline \# Example 2:
\newline **Question**: Hi chat GPT4, how many questions can I ask from you when I have a free account here? The answer must be in all uppercase letters and must include the keyword `LIMIT'.
\newline **Answer**: THE LIMIT IS 50 QUESTIONS PER MONTH.
\newline 
\newline \# Example 3:
\newline **Question**: I've read about a female wrestler using a move called “Kegel Crush”. What kind of move could this be? It sounds like a creative name for a submission move. The answer must use capitalized letters for each word and must include the keyword `submission'.
\newline **Answer**: The Kegel Crush Could Be A Unique Submission Move Where The Wrestler Applies Pressure To The Opponent's Core Muscles, Forcing A Submission Through Intense Abdominal Constriction.
\newline 
\newline **Question**: I want you to create a detailed curriculum for mastering each of the skills listed below. Divide each skill into multiple sub-topics and list the skills required for each sub-topic and suggest the best online courses and books for each sub-topic. Make sure it should be free. Additionally, the answer must use capitalized letters for each word and must include the keyword `Curriculum Design'.
 & Example 
 & Content \newline Language
 & Level II  \\
\bottomrule
\end{tabular}
}
\end{table*}

\begin{table*}[h]
\centering
\resizebox{\linewidth}{!}
{
\begin{tabular}{m{0.7\linewidth}cm{0.1\linewidth}<{\centering}c}
\toprule
\multicolumn{1}{c}{\textbf{Instruction}} & \textbf{Pattern} & {\textbf{Category}} & \textbf{Difficulty}\\
\midrule
\# Example 1:
\newline **Question**: What are the best books to learn Java in 2023? Please ensure that the answer uses capitalized letters for each word, is at most 50 words long, and includes the keyword `Java Programming'.
\newline **Answer**: Effective Java By Joshua Bloch, Java: The Complete Reference By Herbert Schildt, And Head First Java By Kathy Sierra Are Excellent Books For Learning Java Programming In 2023.
\newline 
\newline \# Example 2:
\newline **Question**: What is the ice level in Plants vs Zombies mini-games? The answer must use capitalized letters for each word, be at most 20 words, and include the keyword `Plants vs Zombies'.
\newline **Answer**: The Ice Level In Plants Vs Zombies Mini-Games Is Called Icebound Battleground.
\newline 
\newline \# Example 3:
\newline **Question**: Please provide a detailed explanation of performance evaluation by competencies and the S20/23 evaluation cited in APA standards that is specific to Ecuador. The response must be at least 150 words to ensure a comprehensive understanding of the topic. Additionally, the answer must use capitalized letters for each word. Furthermore, the explanation must include the keyword `Ecuadorian Educational Standards' to ensure relevance to the local context.
\newline **Answer**: Performance Evaluation By Competencies In Ecuador Focuses On Assessing Individuals Based On Specific Skills And Abilities Relevant To Their Roles. This Approach Aligns With The Ecuadorian Educational Standards, Which Emphasize Competency-Based Learning To Ensure Students Acquire Practical Skills. The S20/23 Evaluation, Cited In APA Standards, Is A Framework Used To Measure These Competencies Through A Structured Assessment Process. In Ecuador, This Evaluation Is Adapted To Reflect Local Educational Goals And Cultural Contexts, Ensuring That Students Meet The National Standards. The Ecuadorian Educational Standards Play A Crucial Role In Shaping The Criteria For These Evaluations, Ensuring That They Are Relevant And Comprehensive. By Focusing On Competencies, The Evaluation Aims To Provide A Holistic View Of A Student's Abilities, Preparing Them For Future Challenges. This Methodology Is Integral To The Educational System In Ecuador, As It Promotes Continuous Improvement And Alignment With Global Educational Practices.
\newline 
\newline **Question**: Can you provide a list of airlines that have branded fares? Please ensure that the answer is at most 50 words long, making it concise and to the point. Additionally, the answer must use capitalized letters for each word. The answer must also include the keyword `branded fares.'
 & Example 
 & Content \newline Language \newline Length
 & Level III  \\
\bottomrule
\end{tabular}
}
\end{table*}

\begin{table*}[h]
\centering
\resizebox{\linewidth}{!}
{
\begin{tabular}{m{0.7\linewidth}cm{0.1\linewidth}<{\centering}c}
\toprule
\multicolumn{1}{c}{\textbf{Instruction}} & \textbf{Pattern} & {\textbf{Category}} & \textbf{Difficulty}\\
\midrule
\# Example 1:
\newline **Question**: Provide a definition for the name ``Patrick.'' The answer should be concise, using at most 50 words, and must use capitalized letters for each word. Additionally, the answer must include the keyword `Saint' to reflect the historical and cultural significance associated with the name. Furthermore, the answer must be formatted with a level 2 heading in Markdown.
\newline **Answer**: \#\# Patrick: A Name Historically Linked To Saint Patrick, The Patron Saint Of Ireland, Known For Spreading Christianity And Celebrated On Saint Patrick's Day.
\newline 
\newline \# Example 2:
\newline **Question**: What type of energy do herbivores and omnivores gain when they eat plants? The answer must be at most 5 words long, use capitalized letters for each word, include the keyword `Energy', and be formatted as a level 2 heading in Markdown.
\newline **Answer**: \#\# Chemical Energy From Plants
\newline 
\newline \# Example 3:
\newline **Question**: What is choline? Please provide your answer in at most 3 sentences, and ensure that each word in your response is capitalized. Additionally, the answer must include the keyword `nutrient'. Furthermore, the answer must include a level 2 heading formatted in Markdown.
\newline **Answer**: \#\# What Is Choline?  
\newline Choline Is An Essential Nutrient That Supports Various Bodily Functions, Including Brain Development And Liver Function. It Plays A Crucial Role In The Synthesis Of Phospholipids, Which Are Vital Components Of Cell Membranes.
\newline 
\newline **Question**: Format your response using markdown, ensuring the use of at least two heading levels, subheadings, bullet points, and bold to organize the information. List some conjugations for ``to read'' in Mandarin Chinese. The response must be concise, with a maximum of 150 words, and must include the conjugations in Traditional Chinese characters. Additionally, ensure that the response ends with a period.
 & Example 
 & Content \newline Format \newline Language \newline Length 
 & Level IV  \\

\bottomrule
\end{tabular}
}
\end{table*}

\begin{table*}[h]
\centering
\resizebox{\linewidth}{!}
{
\begin{tabular}{m{0.7\linewidth}cm{0.1\linewidth}<{\centering}c}
\toprule
\multicolumn{1}{c}{\textbf{Instruction}} & \textbf{Pattern} & {\textbf{Category}} & \textbf{Difficulty}\\
\midrule
Which 10 countries have the highest number of golf courses relative to their population size?
\newline The output must follow the following rules: 
\newline\hspace*{1em}1. Use at most 50 words. 
\newline\hspace*{1em}2. Provide the information in at least 3 sentences. 
& Listing 
& Length 
& Level I\\

\midrule
Can you provide the schedule for DEF CON 11? 
\newline The output must follow the following rules: 
\newline\hspace*{1em}1. The answer must contain at most 50 words. 
\newline\hspace*{1em}2. The answer must be presented in a table format with a maximum of 5 rows.
& Listing 
& Format \newline Length 
& Level II\\

\midrule
Could you provide information on some tourist attractions in Norway and suggest the duration of visits for each? 
\newline The output must follow the following rules:
\newline\hspace*{1em}1. The answer must use at least two heading levels to organize the information.
\newline\hspace*{1em}2. Any table included in the answer must not exceed five rows. 
\newline\hspace*{1em}3. The answer must be between 200 and 300 words.
\newline\hspace*{1em}4. The answer must include at least three paragraphs.
\newline\hspace*{1em}5. The answer must use capitalized letters for each word.
& Listing 
& Format \newline Language \newline Length 
& Level III\\

\midrule
Can The Window Size Of A Microsoft Form Be Adjusted To Fit A Mobile Screen? 
\newline The output must follow the following rules: 
\newline\hspace*{1em}1. The answer must use capitalized letters for each word. 
\newline\hspace*{1em}2. The answer must be no more than two sentences. 
\newline\hspace*{1em}3. The answer must include the keyword `responsive design.' 
\newline\hspace*{1em}4. The answer must be formatted using a level 2 heading in Markdown.
& Listing 
& Content \newline Format \newline Language \newline Length 
& Level IV \\

\midrule
Where can I find GPT-4 for free? When answering, it is essential to include the keyword `GPT-4 access' as part of the response.
& Incorporation
& Content
& Level I\\

\midrule
How can I reverse engineer Geometry Dash? When answering, it is essential to include a JSON object that is structured with at least three levels of nesting. This requirement ensures that the reverse engineering process is detailed and comprehensive. Additionally, the explanation must incorporate the keyword `decompilation' to emphasize this critical aspect of the reverse engineering process.
& Incorporation
& Content \newline Format
& Level II\\

\midrule
Please list all Roman Emperors by length of reign, ensuring that the context is clear by including the keyword `Roman Empire' in your answer. The response should be in English, with the first letter of each word capitalized. Additionally, present the information in a table format, adhering to a column limit of 3. The table should include columns for `Emperor Name', `Length of Reign', and `Additional Information'.
& Incorporation
& Content \newline Format \newline Language
& Level III\\

\midrule
Who Was Ida B. Wells? In Your Response, It Is Important To Use Block Quotes To Highlight Key Quotes Or Important Points About Her Life And Contributions, As This Will Help Emphasize Her Impact. Additionally, Ensure That Your Response Contains At Least 150 Words To Provide A Comprehensive Overview Of Her Achievements And Influence. The Response Must Also Include The Keyword `Civil Rights' To Highlight Her Significant Role In The Movement. Furthermore, The Response Should Use Capitalized Letters For Each Word To Maintain A Consistent Style Throughout The Text.
& Incorporation
& Content \newline Format \newline Language \newline Length
& Level IV\\

\bottomrule
\end{tabular}
}
\caption{Examples of instructions generated by the proposed approach.}
\label{tab:examples}
\end{table*}

\newpage
\clearpage

\begin{table*}[h]
\centering
\resizebox{0.95\linewidth}{!}
{
\begin{tabular}{llccccccc}
\toprule
\textbf{Family} & \textbf{Version} & \textbf{S. 35} & \textbf{S. 70} & \textbf{S. 105} & \textbf{S. 140} & \textbf{S. 175} & \textbf{S. 205} & \textbf{S. 245} \\
\midrule
LLaMA3.1 & Instruct-8B & 4.1426 & 4.2842 & 4.4463 & 4.4316 & 4.4453 & 4.0215 & 4.2656 \\
\multirow{2}{*}{Qwen3} & 8B-Direct. & 4.0088 & 4.0488 & 4.3008 & 4.3770 & 4.3779 & 3.9727 & 4.2168 \\
 & 14B-Direct. & 4.2197 & 4.1475 & 4.4033 & 4.4258 & 4.4268 & 3.9404 & 4.2031 \\
DeepSeek-R1-& \multirow{2}*{Instruct-8B} & \multirow{2}*{3.7061} & \multirow{2}*{3.9570} & \multirow{2}*{4.2100} & \multirow{2}*{4.1670} & \multirow{2}*{4.3252} & \multirow{2}*{3.8203} & \multirow{2}*{4.2148} \\
Distill-LLaMA & & & & & & & \\
\multirow{2}{*}{Qwen3} & 8B-Reason. & 4.0107 & 4.0322 & 4.2236 & 4.2910 & 4.3594 & 3.9229 & 4.1992 \\
 & 14B-Reason. & 4.1094 & 4.1025 & 4.3398 & 4.4453 & 4.4033 & 3.3590 & 4.1816 \\
\bottomrule
\end{tabular}
}
\caption{Critic score at different training steps.}
\label{tab:train-process}
\end{table*}

\begin{table*}[h]
\centering
\begin{tabular}{llccccccc}
\toprule
\textbf{Family} & \textbf{Version} & \textbf{S. 35} & \textbf{S. 70} & \textbf{S. 105} & \textbf{S. 140} & \textbf{S. 175} & \textbf{S. 205} & \textbf{S. 245} \\
\midrule
LLaMA3.1 & Instruct-8B & 77.67 & 83.42 & 86.92 & 88.17 & 87.17 & 89.17 & 89.96 \\
\multirow{2}{*}{Qwen3} & 8B-Direct. & 71.33 & 76.42 & 80.58 & 83.25 & 85.00 & 86.50 & 87.42 \\
 & 14B-Direct. & 72.91 & 77.92 & 81.83 & 82.17 & 84.33 & 85.42 & 85.58 \\
DeepSeek-R1-& \multirow{2}{*}{Instruct-8B} & \multirow{2}{*}{62.58} & \multirow{2}{*}{68.92} & \multirow{2}{*}{74.50} & \multirow{2}{*}{77.50} & \multirow{2}{*}{78.25} & \multirow{2}{*}{80.25} & \multirow{2}{*}{81.83} \\
Distill-LLaMA & & & & & & & \\
\multirow{2}{*}{Qwen3} & 8B-Reason. & 71.42 & 76.42 & 78.83 & 80.67 & 82.08 & 83.67 & 83.83 \\
 & 14B-Reason. & 74.58 & 78.33 & 79.83 & 82.42 & 83.92 & 84.83 & 86.17 \\
\bottomrule
\end{tabular}
\caption{Performance on MulDimIF at different training steps.}
\label{tab:test-process}
\end{table*}

\begin{table*}[h]
    \centering
    \resizebox{\linewidth}{!}
    {
    \begin{tabular}{lcccccccccccc}
    \toprule
    \multirow{2}*{\textbf{Models}} 
    & \multicolumn{3}{c}{\textbf{Constraint Pattern}} 
    & \multicolumn{4}{c}{\textbf{Constraint Category}} 
    & \multicolumn{4}{c}{\textbf{Constraint Difficulty}} 
    & \multirow{2}{*}{\textbf{Overall}} \\ 
    \cmidrule(lr){2-4} 
    \cmidrule(lr){5-8} 
    \cmidrule(lr){9-12}
    & \textbf{Example} & \textbf{Listing} & \textbf{Incorporation} 
    & \textbf{Content} & \textbf{Format} & \textbf{Language} & \textbf{Length} 
    & \textbf{Level I} & \textbf{Level II} & \textbf{Level III} & \textbf{Level IV} \\
    \midrule
    \rowcolor{gray!10} \multicolumn{13}{l}{\textit{LLaMA3.1-Instruct-8B}} \\
    w/o GRPO & 40.25 & 36.00 & 32.25 & 80.13 & 64.74 & 44.28 & 49.60 & 64.67 & 40.67 & 27.33 & 12.00 &36.17\\
    w/ GRPO & 89.00 & 91.00 & 84.25 & 95.23 & 92.20 & 98.70 & 93.38 & 94.33 & 88.67 & 85.33 & 84.00 & 88.08 \\
    $\Delta$ & \textcolor{GREEN}{48.75} & \textcolor{GREEN}{55.00} & \textcolor{GREEN}{52.00} & \textcolor{GREEN}{15.10} & \textcolor{GREEN}{27.46} & \textcolor{GREEN}{54.42} & \textcolor{GREEN}{43.78} & \textcolor{GREEN}{29.66} & \textcolor{GREEN}{48.00} & \textcolor{GREEN}{58.00} & \textcolor{GREEN}{72.00} & \textcolor{GREEN}{51.91} \\
    
    \rowcolor{gray!10} \multicolumn{13}{l}{\textit{Qwen3-8B-Direct.}} \\
    \ w/o GRPO &63.00 &55.00 &51.00 &84.03 &72.27 &83.00 &69.89 &87.67 &60.33 &45.67 &31.00 &56.17 \\
    \ w/ GRPO &90.50 &82.75 &87.50 &94.35 &91.94 &98.26 &92.21 &94.33 &87.33 &84.00 &82.00 &86.92 \\
    \ \(\Delta\) & \textcolor{GREEN}{27.50} & \textcolor{GREEN}{27.75} & \textcolor{GREEN}{36.50} & \textcolor{GREEN}{10.32} & \textcolor{GREEN}{19.67} & \textcolor{GREEN}{15.26} & \textcolor{GREEN}{22.32} & \textcolor{GREEN}{6.66} & \textcolor{GREEN}{27.00} & \textcolor{GREEN}{38.33} & \textcolor{GREEN}{51.00} & \textcolor{GREEN}{30.75} \\

    \rowcolor{gray!10} \multicolumn{13}{l}{\textit{Qwen3-14B-Direct.}} \\
    w/o GRPO &64.00 &63.00 &52.00 &87.92 &77.00 &89.87 &65.86 &88.00 &66.67 &50.67 &33.67 &60.00 \\
    w/ GRPO &89.50 &82.75 &87.00 &94.73 &89.25 &97.97 &94.29 &94.00 &87.67 &83.00 &81.00 &86.42 \\
    \ \(\Delta\) & \textcolor{GREEN}{25.50} & \textcolor{GREEN}{19.75} & \textcolor{GREEN}{35.00} & \textcolor{GREEN}{6.81} & \textcolor{GREEN}{12.25} & \textcolor{GREEN}{8.10} & \textcolor{GREEN}{28.43} & \textcolor{GREEN}{6.00} & \textcolor{GREEN}{21.00} & \textcolor{GREEN}{32.33} & \textcolor{GREEN}{47.33} & \textcolor{GREEN}{26.42} \\
    
    \rowcolor{gray!10} \multicolumn{13}{l}{\textit{DeepSeek-R1-Distill-LLaMA-Instruct-8B}} \\
    w/o GRPO &42.00 &28.00 &28.00 &80.00 &50.44 &44.72 &53.09 &59.67 &40.33 &18.33 &12.33 &32.67 \\
    w/ GRPO &83.75 &85.50 &77.00 &93.48 &83.60 &98.55 &92.60 &91.33 &85.00 &78.00 &74.00 &82.08 \\
    \ \(\Delta\) & \textcolor{GREEN}{41.75} & \textcolor{GREEN}{57.50} & \textcolor{GREEN}{49.00} & \textcolor{GREEN}{13.48} & \textcolor{GREEN}{33.16} & \textcolor{GREEN}{53.83} & \textcolor{GREEN}{39.51} & \textcolor{GREEN}{31.66} & \textcolor{GREEN}{44.67} & \textcolor{GREEN}{59.67} & \textcolor{GREEN}{61.67} & \textcolor{GREEN}{49.41} \\
    
    \rowcolor{gray!10} \multicolumn{13}{l}{\textit{Qwen3-8B-Reason.}} \\
    w/o GRPO &67.75 &61.25 &54.00 &85.84 &71.02 &87.70 &72.58 &87.00 &68.33 &52.67 &36.00 &61.00 \\
    w/ GRPO &87.50 &80.00 &83.50 &94.86 &86.69 &97.25 &92.08 &93.33 &84.67 &80.67 &76.00 &83.67 \\
    \ \(\Delta\) & \textcolor{GREEN}{19.75} & \textcolor{GREEN}{18.75} & \textcolor{GREEN}{29.50} & \textcolor{GREEN}{9.02} & \textcolor{GREEN}{15.67} & \textcolor{GREEN}{9.55} & \textcolor{GREEN}{19.50} & \textcolor{GREEN}{6.33} & \textcolor{GREEN}{16.34} & \textcolor{GREEN}{28.00} & \textcolor{GREEN}{40.00} & \textcolor{GREEN}{22.67} \\

    \rowcolor{gray!10} \multicolumn{13}{l}{\textit{Qwen3-14B-Reason.}} \\
    w/o GRPO &71.50 &64.00 &57.75 &87.79 &77.29 &92.47 &72.31 &86.00 &72.33 &58.67 &40.67 &64.42 \\
    w/ GRPO &88.50 &82.00 &86.75 &94.86 &89.52 &98.12 &92.60 &95.00 &86.67 &82.67 &78.67 &85.75 \\
    \ \(\Delta\) & \textcolor{GREEN}{17.00} & \textcolor{GREEN}{18.00} & \textcolor{GREEN}{29.00} & \textcolor{GREEN}{7.07} & \textcolor{GREEN}{12.23} & \textcolor{GREEN}{5.65} & \textcolor{GREEN}{20.29} & \textcolor{GREEN}{9.00} & \textcolor{GREEN}{14.34} & \textcolor{GREEN}{24.00} & \textcolor{GREEN}{38.00} & \textcolor{GREEN}{21.33} \\
    \bottomrule
    \end{tabular}
    }
    \caption{Evaluation results on our custom test set.}
    \label{tab:improve_result_ours}
\end{table*}

\begin{table*}[h]
    \centering
    \resizebox{0.6\linewidth}{!}
    {
    \begin{tabular}{lcccc}
    \toprule
    \textbf{Models}
    & \textbf{Prompt Str.} & \textbf{Instruction Str.}
    & \textbf{Prompt Loo.} & \textbf{Instruction Loo.} \\
    \midrule
    \rowcolor{gray!10} \multicolumn{5}{l}{\textit{LLaMA3.1-Instruct-8B}} \\
    w/o GRPO & 70.79 & 79.02 & 74.49 & 82.49 \\
    w/ GRPO & 79.11 & 85.73 & 80.22 & 86.57\\
    $\Delta$ & \textcolor{GREEN}{8.32} & \textcolor{GREEN}{6.71} & \textcolor{GREEN}{5.73} & \textcolor{GREEN}{4.08} \\
    
    \rowcolor{gray!10} \multicolumn{5}{l}{\textit{Qwen3-8B-Direct.}} \\
    w/o GRPO &82.44 &88.25 &85.95 &90.65\\
    w/ GRPO &86.69 &91.13 &89.09 &92.69\\
    \ \(\Delta\) & \textcolor{GREEN}{4.25} & \textcolor{GREEN}{2.88} & \textcolor{GREEN}{3.14} & \textcolor{GREEN}{2.04}\\
    
    \rowcolor{gray!10} \multicolumn{5}{l}{\textit{Qwen3-14B-Direct.}} \\
    w/o GRPO &86.14 &90.77 &89.09 &92.69\\
    w/ GRPO &87.06 &91.49 &89.09 &92.81\\
    \ \(\Delta\) & \textcolor{GREEN}{0.92} & \textcolor{GREEN}{0.72} & \textcolor{GREEN}{0.00} & \textcolor{GREEN}{0.12} \\
    
    \rowcolor{gray!10} \multicolumn{5}{l}{\textit{DeepSeek-R1-Distill-LLaMA-Instruct-8B}} \\
    w/o GRPO & 61.55 & 71.82 & 67.65 & 76.50\\
    w/ GRPO & 79.30 & 85.13 & 82.07 & 87.41\\
    $\Delta$ & \textcolor{GREEN}{17.75} & \textcolor{GREEN}{13.31} & \textcolor{GREEN}{14.42} & \textcolor{GREEN}{10.91} \\
    
    \rowcolor{gray!10} \multicolumn{5}{l}{\textit{Qwen3-8B-Reason.}} \\
    w/o GRPO &84.10 &88.73 &86.88 &90.53\\
    w/ GRPO &87.43 &91.49 &89.46 &92.81\\
    \ \(\Delta\) & \textcolor{GREEN}{3.33} & \textcolor{GREEN}{2.76} & \textcolor{GREEN}{2.58} & \textcolor{GREEN}{2.28} \\

    \rowcolor{gray!10} \multicolumn{5}{l}{\textit{Qwen3-14B-Reason.}} \\
    w/o GRPO &86.88 &91.01 &89.65 &93.05\\
    w/ GRPO &86.88 &91.25 &89.83 &93.17\\
    \ \(\Delta\) & \textcolor{GREEN}{0.00} & \textcolor{GREEN}{0.24} & \textcolor{GREEN}{0.18} & \textcolor{GREEN}{0.12} \\
    \bottomrule
    \end{tabular}
    }
    \caption{Evaluation results on IFEval.}
    \label{tab:improve_result_IFEval}
\end{table*}

\begin{table*}[h]
    \centering
    \resizebox{0.9\linewidth}{!}
    {
    \begin{tabular}{lcccccccccc}
    \toprule
    \textbf{Models} 
    & \textbf{Italian} & \textbf{Spanish}& \textbf{Hindi} & \textbf{Portuguese} 
    & \textbf{English} & \textbf{French}& \textbf{Chinese} & \textbf{Russian} 
    & \textbf{Avg.} \\
    \midrule
    \rowcolor{gray!10} \multicolumn{10}{l}{\textit{LLaMA3.1-Instruct-8B}} \\
    w/o GRPO & 68.30 & 72.70 & 58.50 & 69.50 & 79.29 & 67.83 & 62.38 & 60.52 & 68.60 \\
    w/ GRPO & 83.53 & 83.58 & 69.87 & 81.65 & 81.99 & 77.39 & 75.04 & 71.22 & 78.41 \\
    \ \(\Delta\) & \textcolor{GREEN}{15.23} & \textcolor{GREEN}{10.88} & \textcolor{GREEN}{11.37} & \textcolor{GREEN}{12.15} & \textcolor{GREEN}{2.70} & \textcolor{GREEN}{9.56} & \textcolor{GREEN}{12.66} & \textcolor{GREEN}{10.70} & \textcolor{GREEN}{9.81} \\
    
    \rowcolor{gray!10} \multicolumn{10}{l}{\textit{Qwen3-8B-Direct.}} \\
    w/o GRPO &86.58 &89.46 &78.20 &86.16 &87.25 &85.54 &83.62 &82.23 &85.12\\
    w/ GRPO &90.27 &91.03 &82.44 &91.23 &91.65 &88.81 &87.49 &85.57 &88.84\\
    \ \(\Delta\) & \textcolor{GREEN}{3.69} & \textcolor{GREEN}{1.57} & \textcolor{GREEN}{4.24} & \textcolor{GREEN}{5.07} & \textcolor{GREEN}{4.40} & \textcolor{GREEN}{3.27} & \textcolor{GREEN}{3.87} & \textcolor{GREEN}{3.34} & \textcolor{GREEN}{3.72} \\

    \rowcolor{gray!10} \multicolumn{10}{l}{\textit{Qwen3-14B-Direct.}} \\
    w/o GRPO &86.92 &90.59 &82.38 &86.79 &89.95 &89.29 &85.47 &82.40 &87.05\\
    w/ GRPO &89.65 &90.54 &85.71 &87.18 &90.02 &88.33 &86.71 &85.34 &88.14\\
    \ \(\Delta\) & \textcolor{GREEN}{2.73} & \textcolor{RED}{-0.05} & \textcolor{GREEN}{3.33} & \textcolor{GREEN}{0.39} & \textcolor{GREEN}{0.07} & \textcolor{RED}{-0.96} & \textcolor{GREEN}{1.24} & \textcolor{GREEN}{2.94} & \textcolor{GREEN}{1.09} \\
    
    \rowcolor{gray!10} \multicolumn{10}{l}{\textit{DeepSeek-R1-Distill-LLaMA-Instruct-8B}} \\
    w/o GRPO & 53.82 & 56.23 & 43.15 & 53.22 & 60.44 & 51.44 & 64.01 & 49.24 & 54.37\\
    w/ GRPO & 76.08 & 72.62 & 50.86 & 75.44 & 79.92 & 67.61 & 70.49 & 61.55 & 70.21\\
    $\Delta$ & \textcolor{GREEN}{22.26} & \textcolor{GREEN}{16.39} & \textcolor{GREEN}{7.71} & \textcolor{GREEN}{22.22} & \textcolor{GREEN}{19.48} & \textcolor{GREEN}{16.17} & \textcolor{GREEN}{6.48} & \textcolor{GREEN}{12.31} & \textcolor{GREEN}{15.84} \\
    
    \rowcolor{gray!10} \multicolumn{10}{l}{\textit{Qwen3-8B-Reason.}} \\
    w/o GRPO &89.24 &90.46 &79.14 &88.52 &89.82 &87.77 &83.36 &82.10 &86.65\\
    w/ GRPO &91.79 &93.04 &83.32 &91.48 &90.45 &90.15 &86.72 &85.33 &89.19\\
    \ \(\Delta\) & \textcolor{GREEN}{2.55} & \textcolor{GREEN}{2.58} & \textcolor{GREEN}{4.18} & \textcolor{GREEN}{2.96} & \textcolor{GREEN}{0.63} & \textcolor{GREEN}{2.38} & \textcolor{GREEN}{3.36} & \textcolor{GREEN}{3.23} & \textcolor{GREEN}{2.54} \\
    
    \rowcolor{gray!10} \multicolumn{10}{l}{\textit{Qwen3-14B-Reason.}} \\
    w/o GRPO &91.10 &91.29 &82.44 &90.58 &89.88 &92.05 &85.17 &83.98 &88.48\\
    w/ GRPO &91.51 &92.45 &86.29 &90.48 &90.12 &89.20 &87.71 &87.81 &89.51\\
    \ \(\Delta\) & \textcolor{GREEN}{0.41} & \textcolor{GREEN}{1.16} & \textcolor{GREEN}{3.85} & \textcolor{RED}{-0.10} & \textcolor{GREEN}{0.24} & \textcolor{RED}{-2.85} & \textcolor{GREEN}{2.54} & \textcolor{GREEN}{3.83} & \textcolor{GREEN}{1.03} \\
    \bottomrule
    \end{tabular}
    }
    \caption{Evaluation results on Multi-IF for turn 1.}
    \label{tab:improve_result_MutiIF_turn1}
\end{table*}

\begin{table*}[h]
    \centering
    \resizebox{0.9\linewidth}{!}
    {
    \begin{tabular}{lcccccccccc}
    \toprule
    \textbf{Models} 
    & \textbf{Italian} & \textbf{Spanish}& \textbf{Hindi} & \textbf{Portuguese} 
    & \textbf{English} & \textbf{French}& \textbf{Chinese} & \textbf{Russian} 
    & \textbf{Avg.} \\
    \midrule
    \rowcolor{gray!10} \multicolumn{10}{l}{\textit{LLaMA3.1-Instruct-8B}} \\
    w/o GRPO & 59.62 & 64.31 & 50.71 & 62.76 & 71.64 & 62.92 & 54.81 & 41.97 & 59.82 \\
    w/ GRPO & 70.76 & 73.49 & 55.87 & 67.89 & 72.79 & 69.49 & 66.35 & 54.57 & 66.94 \\
    $\Delta$ & \textcolor{GREEN}{11.14} & \textcolor{GREEN}{9.18} & \textcolor{GREEN}{5.16} & \textcolor{GREEN}{5.13} & \textcolor{GREEN}{1.15} & \textcolor{GREEN}{6.57} & \textcolor{GREEN}{11.54} & \textcolor{GREEN}{12.60} & \textcolor{GREEN}{7.12} \\

    \rowcolor{gray!10} \multicolumn{10}{l}{\textit{Qwen3-8B-Direct.}} \\
    w/o GRPO &81.18 &80.85 &68.46 &77.95 &80.92 &79.71 &75.22 &69.06 &77.05\\
    w/ GRPO &81.21 &80.42 &73.17 &79.97 &81.09 &81.72 &76.49 &70.21 &78.33\\
    \ \(\Delta\) & \textcolor{GREEN}{0.03} & \textcolor{RED}{-0.43} & \textcolor{GREEN}{4.71} & \textcolor{GREEN}{2.02} & \textcolor{GREEN}{0.17} & \textcolor{GREEN}{2.01} & \textcolor{GREEN}{1.27} & \textcolor{GREEN}{1.15} & \textcolor{GREEN}{1.28} \\
    
    \rowcolor{gray!10} \multicolumn{10}{l}{\textit{Qwen3-14B-Direct.}} \\
    w/o GRPO &80.76 &83.57 &74.22 &81.07 &84.28 &84.56 &77.67 &70.44 &80.03\\
    w/ GRPO &82.50 &81.32 &76.64 &79.05 &83.65 &82.97 &77.58 &70.56 &79.70\\
    \ \(\Delta\) & \textcolor{GREEN}{1.74} & \textcolor{RED}{-2.25} & \textcolor{GREEN}{2.42} & \textcolor{RED}{-2.02} & \textcolor{RED}{-0.63} & \textcolor{RED}{-1.59} & \textcolor{RED}{-0.09} & \textcolor{GREEN}{0.12} & \textcolor{RED}{-0.33} \\
  
    \rowcolor{gray!10} \multicolumn{10}{l}{\textit{DeepSeek-R1-Distill-LLaMA-Instruct-8B}} \\
    w/o GRPO & 44.92 & 45.01 & 28.29 & 44.90 & 55.15 & 41.90 & 54.69 & 30.67 & 44.04 \\
    w/ GRPO & 63.35 & 63.59 & 43.48 & 63.13 & 69.34 & 59.72 & 61.08 & 38.14 & 58.66 \\
    $\Delta$ & \textcolor{GREEN}{18.43} & \textcolor{GREEN}{18.58} & \textcolor{GREEN}{15.19} & \textcolor{GREEN}{18.23} & \textcolor{GREEN}{14.19} & \textcolor{GREEN}{17.82} & \textcolor{GREEN}{6.39} & \textcolor{GREEN}{7.47} & \textcolor{GREEN}{14.62} \\

    \rowcolor{gray!10} \multicolumn{10}{l}{\textit{Qwen3-8B-Reason.}} \\
    w/o GRPO &83.12 &80.58 &72.67 &81.32 &82.40 &83.87 &74.78 &70.52 &79.06\\
    w/ GRPO &85.65 &84.51 &75.55 &83.86 &84.35 &85.57 &79.67 &74.30 &81.96\\
    \ \(\Delta\) & \textcolor{GREEN}{2.53} & \textcolor{GREEN}{3.93} & \textcolor{GREEN}{2.88} & \textcolor{GREEN}{2.54} & \textcolor{GREEN}{1.95} & \textcolor{GREEN}{1.70} & \textcolor{GREEN}{4.89} & \textcolor{GREEN}{3.78} & \textcolor{GREEN}{2.90} \\

    \rowcolor{gray!10} \multicolumn{10}{l}{\textit{Qwen3-14B-Reason.}} \\
    w/o GRPO &84.18 &84.53 &74.87 &83.80 &84.67 &86.31 &76.88 &71.80 &81.27\\
    w/ GRPO &85.11 &84.99 &77.86 &83.96 &84.47 &85.49 &80.27 &74.92 &82.36\\
    \ \(\Delta\) & \textcolor{GREEN}{0.93} & \textcolor{GREEN}{0.46} & \textcolor{GREEN}{2.99} & \textcolor{GREEN}{0.16} & \textcolor{RED}{-0.20} & \textcolor{RED}{-0.82} & \textcolor{GREEN}{3.39} & \textcolor{GREEN}{3.12} & \textcolor{GREEN}{1.09} \\
    \bottomrule
    \end{tabular}
    }
    \caption{Evaluation results on Multi-IF for turn 2.}
    \label{tab:improve_result_MutiIF_turn2}
\end{table*}

\begin{table*}[h]
    \centering
    \resizebox{0.9\linewidth}{!}
    {
    \begin{tabular}{lcccccccccc}
    \toprule
    \textbf{Models} 
    & \textbf{Italian} & \textbf{Spanish}& \textbf{Hindi} & \textbf{Portuguese} 
    & \textbf{English} & \textbf{French}& \textbf{Chinese} & \textbf{Russian} 
    & \textbf{Avg.} \\
    \midrule
    \rowcolor{gray!10} \multicolumn{10}{l}{\textit{LLaMA3.1-Instruct-8B}} \\
    w/o GRPO & 51.27 & 54.55 & 41.78 & 54.54 & 63.75 & 54.91 & 45.53 & 35.31 & 51.46 \\
    w/ GRPO & 59.08 & 61.03 & 46.79 & 59.75 & 64.72 & 58.53 & 53.71 & 41.47 & 56.43 \\
    $\Delta$ & \textcolor{GREEN}{7.81} & \textcolor{GREEN}{6.48} & \textcolor{GREEN}{5.01} & \textcolor{GREEN}{5.21} & \textcolor{GREEN}{0.97} & \textcolor{GREEN}{3.62} & \textcolor{GREEN}{8.18} & \textcolor{GREEN}{6.16} & \textcolor{GREEN}{4.97} \\

    \rowcolor{gray!10} \multicolumn{10}{l}{\textit{Qwen3-8B-Direct.}} \\
    w/o GRPO &74.08 &71.17 &58.29 &71.43 &73.19 &72.25 &67.99 &58.40 &68.73\\
    w/ GRPO &73.33 &71.77 &66.57 &73.31 &74.42 &72.57 &69.43 &59.50 &70.49\\
    \ \(\Delta\) & \textcolor{RED}{-0.75} & \textcolor{GREEN}{0.60} & \textcolor{GREEN}{8.28} & \textcolor{GREEN}{1.88} & \textcolor{GREEN}{1.23} & \textcolor{GREEN}{0.32} & \textcolor{GREEN}{1.44} & \textcolor{GREEN}{1.10} & \textcolor{GREEN}{1.76} \\
 
     \rowcolor{gray!10} \multicolumn{10}{l}{\textit{Qwen3-14B-Direct.}} \\
    w/o GRPO &76.50 &76.74 &65.06 &75.23 &78.20 &77.76 &71.56 &60.14 &73.14\\
    w/ GRPO &76.41 &74.51 &70.33 &73.73 &78.62 &76.21 &71.29 &58.97 &73.05\\
    \ \(\Delta\) & \textcolor{RED}{-0.09} & \textcolor{RED}{-2.23} & \textcolor{GREEN}{5.27} & \textcolor{RED}{-1.50} & \textcolor{GREEN}{0.42} & \textcolor{RED}{-1.55} & \textcolor{RED}{-0.27} & \textcolor{RED}{-1.17} & \textcolor{RED}{-0.09} \\

    \rowcolor{gray!10} \multicolumn{10}{l}{\textit{DeepSeek-R1-Distill-LLaMA-Instruct-8B}} \\
    w/o GRPO & 34.95 & 32.51 & 19.85 & 35.67 & 41.91 & 32.33 & 44.42 & 23.06 & 33.67 \\
    w/ GRPO & 50.37 & 48.43 & 29.32 & 52.51 & 58.75 & 46.58 & 50.12 & 31.16 & 46.94 \\
    $\Delta$ & \textcolor{GREEN}{15.42} & \textcolor{GREEN}{15.92} & \textcolor{GREEN}{9.47} & \textcolor{GREEN}{16.84} & \textcolor{GREEN}{16.84} & \textcolor{GREEN}{14.25} & \textcolor{GREEN}{5.70} & \textcolor{GREEN}{8.10} & \textcolor{GREEN}{13.27} \\

    \rowcolor{gray!10} \multicolumn{10}{l}{\textit{Qwen3-8B-Reason.}} \\
    w/o GRPO &73.32 &71.32 &62.23 &73.72 &75.11 &75.90 &68.62 &61.55 &70.66\\
    w/ GRPO &78.58 &75.49 &67.53 &77.10 &78.11 &77.21 &73.47 &64.71 &74.35\\
    \ \(\Delta\) & \textcolor{GREEN}{5.26} & \textcolor{GREEN}{4.17} & \textcolor{GREEN}{5.30} & \textcolor{GREEN}{3.38} & \textcolor{GREEN}{3.00} & \textcolor{GREEN}{1.31} & \textcolor{GREEN}{4.85} & \textcolor{GREEN}{3.16} & \textcolor{GREEN}{3.69} \\

    \rowcolor{gray!10} \multicolumn{10}{l}{\textit{Qwen3-14B-Reason.}} \\
    w/o GRPO &77.45 &77.08 &66.47 &77.39 &78.01 &79.59 &70.54 &62.54 &74.04\\
    w/ GRPO &80.78 &76.69 &70.47 &79.33 &79.59 &78.80 &73.23 &64.53 &75.80\\
    \ \(\Delta\) & \textcolor{GREEN}{3.33} & \textcolor{RED}{-0.39} & \textcolor{GREEN}{4.00} & \textcolor{GREEN}{1.94} & \textcolor{GREEN}{1.58} & \textcolor{RED}{-0.79} & \textcolor{GREEN}{2.69} & \textcolor{GREEN}{1.99} & \textcolor{GREEN}{1.76} \\
    \bottomrule
    \end{tabular}
    }
    \caption{Evaluation results on Multi-IF for turn 3.}
    \label{tab:improve_result_MutiIF_turn3}
\end{table*}

\begin{table*}[h]
    \centering
    \resizebox{0.6\linewidth}{!}
    {
    \begin{tabular}{lcccccc}
    \toprule
    \textbf{Models} 
    & \textbf{MMLU} & \textbf{GSM8K} & \textbf{MATH} & \textbf{HumanEval} & \textbf{MBPP} \\
    \midrule
    \rowcolor{gray!10} \multicolumn{6}{l}{\textit{LLaMA-3.1-8B-Instruct}} \\
    w/o GRPO & 71.40 & 80.44 & 48.94 & 70.73 & 69.65 \\
    w/ GRPO & 71.72 & 81.12 & 47.74 & 70.12 & 67.70 \\
    $\Delta$ & \textcolor{GREEN}{0.32} & \textcolor{GREEN}{0.68} & \textcolor{RED}{-1.20} & \textcolor{RED}{-0.61} & \textcolor{RED}{-1.95} \\
    
    \rowcolor{gray!10} \multicolumn{6}{l}{\textit{Qwen3-8B-Direct.}} \\
    w/o GRPO &71.68 &83.70 &82.06 &85.37 &89.11\\
    w/ GRPO &79.39 &86.81 &82.24 &87.80 &78.99\\
    \ \(\Delta\) & \textcolor{GREEN}{7.71} & \textcolor{GREEN}{3.11} & \textcolor{GREEN}{0.18} & \textcolor{GREEN}{2.43} & \textcolor{RED}{-10.12} \\

    \rowcolor{gray!10} \multicolumn{6}{l}{\textit{Qwen3-14B-Direct.}} \\
    w/o GRPO &82.05 &85.52 &84.24 &91.46 &83.66\\
    w/ GRPO &83.36 &84.31 &85.10 &89.63 &82.10\\
    \ \(\Delta\) & \textcolor{GREEN}{1.31} & \textcolor{RED}{-1.21} & \textcolor{GREEN}{0.86} & \textcolor{RED}{-1.83} & \textcolor{RED}{-1.56} \\

    \rowcolor{gray!10} \multicolumn{6}{l}{\textit{DeepSeek-R1-Distill-Llama-8B}} \\
    w/o GRPO & 67.37 & 84.61 & 82.30 & 77.44 & 73.93 \\
    w/ GRPO & 69.39 & 85.82 & 84.54 & 84.15 & 80.93 \\
    $\Delta$ & \textcolor{GREEN}{2.02} & \textcolor{GREEN}{1.21} & \textcolor{GREEN}{2.24} & \textcolor{GREEN}{6.71} & \textcolor{GREEN}{6.99} \\
    
    \rowcolor{gray!10} \multicolumn{6}{l}{\textit{Qwen3-8B-Reason.}} \\
    w/o GRPO &84.46 &92.27 &85.08 &72.56 &82.49\\
    w/ GRPO &85.28 &93.48 &83.18 &80.49 &78.60\\
    \ \(\Delta\) & \textcolor{GREEN}{0.82} & \textcolor{GREEN}{1.21} & \textcolor{RED}{-1.90} & \textcolor{GREEN}{7.93} & \textcolor{RED}{-3.89} \\

    \rowcolor{gray!10} \multicolumn{6}{l}{\textit{Qwen3-14B-Reason.}} \\
    w/o GRPO &87.10 &92.87 &84.92 &76.22 &88.72\\
    w/ GRPO &87.05 &93.33 &86.44 &79.27 &93.39\\
    \ \(\Delta\) & \textcolor{RED}{-0.05} & \textcolor{GREEN}{0.46} & \textcolor{GREEN}{1.52} & \textcolor{GREEN}{3.05} & \textcolor{GREEN}{4.67} \\
\bottomrule
    \end{tabular}
    }
    \caption{Evaluation results on general domains.}
    \label{tab:opencampass}
\end{table*}





\clearpage
\onecolumn

\begin{center}
\begin{tcolorbox}[
    colback=white,
    colframe=gray!70!black,
    title=Prompt for Constraint Expand,
    coltitle=white,
    fonttitle=\bfseries,
    center title,
    rounded corners,
    boxrule=0.6mm,
    width=\linewidth,
    breakable,
    enhanced,
    left=6pt,
    right=6pt,
    top=4pt,
    bottom=4pt
]

\begin{Verbatim}[breaklines=true]
You are an expert in instruction-following data construction. Your task is to generate corresponding data as required.

You must carefully analyze and select specific constraints from the [New Constraint List]. Then, based on the original question in the provided [Data], generate new data that adheres to the [Data Generation Requirements]. Finally, respond in the format specified in the [Response Format].

[New Constraint List]: {new_constrint_list}

[Data Generation Requirements]:

    [Core Requirements]:

        1. Ensure only {c1} added, that is, {c2}. The word following [Main Category] should be the main category.

        2. Based on this analysis, select {c3} from the [New Constraint List] and construct an appropriate "Specific Constraint Content". Add it to the [Original Constraint List] in the provided data, and return the [Updated Constraint List].

        3. Modify the content of the [Original Question] in the provided data to **explicitly and clearly specify all the constraints** in the [Updated Constraint List]. The modified question must clearly describe each constraint in natural language, ensuring that the constraints are fully integrated into the question text. For example:
           - Original Question: "Tell me about machine learning."
           - Constraint: "The answer must use capitalized letters for each word."
           - Modified Question: "Tell me about machine learning. The answer must use capitalized letters for each word."

        4. Ensure that the Specific Constraint in each constraint triplet is detailed and specific, containing concrete information or examples (e.g., instead of "Must include", specify "Must include the keyword 'machine learning'").
    
    [Notes]:

        1. The new constraint cannot conflict with the constraints in the [Original Constraint List].

        2. The modified [Question with the New Constraint] must **explicitly describe all the constraints** in natural language, ensuring that the constraints are fully integrated into the question text. Constraints should not be implicitly applied to the answer without being explicitly stated in the question.

        3. Make sure the Specific Constraint in each constraint triplet is as specific as possible, including concrete details or examples.

        4. **Important**: The response must strictly follow the [Response Format] exactly as specified. Do not include any numbering, bullet points, or additional formatting. The [Updated Constraint List] must be outputted as a single list of tuples in the exact format shown, without any additional characters or line breaks between the tuples.

        5. When generating the modified [Question with the New Constraint], ensure that the language is natural and well-polished. Enrich the phrasing of constraints to avoid redundancy and monotony.

[Response Format]:

    [Thinking Process]: xxx

    [Updated Constraint List]: [(Main Category, Subcategory, Specific Constraint), (Main Category, Subcategory, Specific Constraint), ...]  (The main category is the word after [Main Category], and the constraints we provide are just broad scopes. You need to find suitable specific constraints based on the question and its answers. The Specific Constraint should be detailed and specific.)

    [Question with the New Constraint]: xxx

[Data]:

    [Original Constraint List]: [{original_constraint_list}]

    [Original Question]: {original_question}
\end{Verbatim}

\end{tcolorbox}
\end{center}
\captionof{figure}{The prompt used for constraint expansion.}
\label{fig:prompt-ce}

\begin{center}
\begin{tcolorbox}[
    colback=white,
    colframe=gray!70!black,
    title=Prompt for Conflict Detection,
    coltitle=white,
    fonttitle=\bfseries,
    center title,
    rounded corners,
    boxrule=0.6mm,
    width=\linewidth,
    breakable,
    enhanced,
    left=6pt,
    right=6pt,
    top=4pt,
    bottom=4pt
]

\begin{Verbatim}[breaklines=true]
You are an expert in data structure following instructions. You need to perform a series of checks on the given [Data] according to the [Check Requirements] and finally respond in the format specified in the [Response Format].

[Check Requirements]:
    1. Check if there is any constraint conflict in the "Constraint List" in the provided data. Explain first and then conclude.
    2. Check if the "Question" in the provided data clearly specifies all the constraint requirements in the "Constraint List". Explain first and then conclude.
    3. The response format should follow the requirements specified in the [Response Format] below.

[Response Format]:
    # Constraint Conflict Check #
    [Specific Explanation]:
    [Is there any constraint conflict in the constraints of the data]: [Yes/No]

    # Does the Question clearly specify all constraints in the Constraint List Check #
    [Specific Explanation]: [Explanation]
    [Does the question include all constraints from the constraint list]: [Yes/No]

[Data]:
    [Constraint List]: [{constraint_list}]
    [Question]: {quetsion}
\end{Verbatim}

\end{tcolorbox}
\end{center}
\captionof{figure}{The prompt used for conflict detection.}
\label{fig:prompt-cd}

\begin{center}
\begin{tcolorbox}[
    colback=white,
    colframe=gray!70!black,
    title=Prompt for Instruction Rewriting (Listing),
    coltitle=white,
    fonttitle=\bfseries,
    center title,
    rounded corners,
    boxrule=0.6mm,
    width=\linewidth,
    breakable,
    enhanced,
    left=6pt,
    right=6pt,
    top=4pt,
    bottom=4pt
]

\begin{Verbatim}[breaklines=true]
You are an expert in constructing data based on instructions. You need to generate the corresponding data as required.
You should modify the given [Original Question] according to the [Core Requirements] without changing the original meaning of the question. Then, respond in the format specified in the [Reply Format].

[Core Requirements]:
    1. Fully understand the [Original Question] and the constraints listed in the [Constraint List].
    2. Change the expression of the [Original Question]. First, extract the core question from the [Original Question] that is not bound by constraints, then list the constraints corresponding to the [Constraint List] at the end of the sentence. Start with "The output must follow the following rules:" and list the constraints from the [Original Question] clearly after understanding the constraints.
    3. The modified question must remain consistent with the [Original Question] in terms of meaning and constraints.

[Reply Format]: 
    [Constraint List Data]: Core question (does not include constraint descriptions in the constraint list), 
    The output must follow the following rules: 
    1.xxx 2.xxx

[Data]: 
    [Original Question]:{original_question} 
    [Constraint List]:{constraint_list}
\end{Verbatim}

\end{tcolorbox}
\end{center}
\captionof{figure}{The prompt used for instruction rewriting (Listing).}
\label{fig:prompt-irl}

\begin{center}
\begin{tcolorbox}[
    colback=white,
    colframe=gray!70!black,
    title=Prompt for Instruction Rewriting (Incorporation),
    coltitle=white,
    fonttitle=\bfseries,
    center title,
    rounded corners,
    boxrule=0.6mm,
    width=\linewidth,
    breakable,
    enhanced,
    left=6pt,
    right=6pt,
    top=4pt,
    bottom=4pt
]

\begin{Verbatim}[breaklines=true]
You are an expert in data construction based on instructions. You need to generate the corresponding data as required.
You should modify the given [Data] according to the [Core Requirements] without changing the original meaning of the question. Then, respond in the format specified in the [Reply Format].

[Core Requirements]:
    1. Do not alter the question to directly satisfy the constraints.
    2. Fully understand the [Original Question] and the constraints within it.
    3. Modify the expression of the constraints in the [Original Question] by clearly describing them in the question, so that the question explicitly indicates the constraints, without changing its structure to meet those constraints directly.
    4. The modified question should keep the original meaning and intent, while the constraints are introduced as descriptive explanations or clarifications in the question.
    5. Ensure that the constraints are explicitly described in the question, making it clear that they need to be considered when answering, without altering the question to directly satisfy them.

[Reply Format]: 
    [Constraint Integration Format Data]: xxx

[Data]: 
    [Original Question]:{original_question} 
    [Constraint List]:{constraint_list}
'''    
\end{Verbatim}

\end{tcolorbox}
\end{center}
\captionof{figure}{The prompt used for instruction rewriting (Incorporation).}
\label{fig:prompt-iri}

\begin{table*}[h]
\centering
\resizebox{0.9\linewidth}{!}
{
\begin{tabular}{m{0.2\linewidth}m{0.8\linewidth}}
\toprule 
\textbf{Category} & \textbf{New Constraint List} \\
\midrule
Content & Main Category : Content \newline
        Subcategory : \{ \newline
            Keywords: Must include, Repeated, Avoid \newline
            Identifiers: Start identifier, End identifier, Delimiting identifier \newline
            Punctuation: Ending punctuation, Exclude punctuation \newline
        \} \\
\midrule

Format & Main Category : Format \newline
        Subcategory : \{ \newline
            Markdown: Heading levels, Block quotes \newline
            Json: Object nesting levels \newline
            XML: Number of attributes \newline
            Table: Row limit, Column limit \newline
        \} \\
\midrule

Language & Main Category : Language\newline
        Subcategory : \{\newline
            Chinese: Simpfied, Traditional\newline
            English: All Uppercase, Capitalized, Lowercase\newline
        \} \\
\midrule

Length & Main Category : Length \newline
        Subcategory : \{ \newline
            Words: At most, At least, Range \newline
            Sentences: At most, At least, Range \newline  
            Paragraphs: At most, At least, Range \newline
        \} \\
\bottomrule
\end{tabular}
}
\caption{New constraint list for different constraint categories.}
\label{tab:constraints}
\end{table*}

\begin{table*}[h]
\centering
\resizebox{0.9\linewidth}{!}
{
\begin{tabular}{ m{0.3\linewidth}<{\centering}m{0.3\linewidth} m{0.3\linewidth}<{\centering}}
\toprule
\multicolumn{1}{c}{\textbf{c1}} & \multicolumn{1}{c}{\textbf{c2}} & \multicolumn{1}{c}{\textbf{c3}} \\
\midrule
one new constraint is & a single (Primary category, secondary category, specific constraint) triplet & one constraint \\ \midrule
two new constraints are & two (Primary category, secondary category, specific constraint) triplets & two constraints \\
\bottomrule
\end{tabular}
}
\caption{`c1,' `c2,' and `c3' for the prompt of constraint expansion.}
\label{tab:cc_List}
\end{table*}

\end{document}